\documentclass{article} 
\usepackage{iclr2025_conference,times}

\usepackage{amsmath,amsfonts,bm}

\def\eqref#1{equation~\ref{#1}}

\def\1{\bm{1}}

\DeclareMathAlphabet{\mathsfit}{\encodingdefault}{\sfdefault}{m}{sl}
\SetMathAlphabet{\mathsfit}{bold}{\encodingdefault}{\sfdefault}{bx}{n}

\usepackage{hyperref}
\usepackage{url}
\usepackage{graphicx}
\usepackage{booktabs}
\usepackage{multirow}
\usepackage{wrapfig}
\usepackage{floatrow}
\usepackage{caption}

\newfloatcommand{capbtabbox}{table}[][\FBwidth]

\title{MVTokenFlow: High-quality 4D Content Generation using Multiview Token Flow}

\author{
Hanzhuo Huang\textsuperscript{1}\thanks{Equal contribution.},~~
Yuan Liu\textsuperscript{2$\ast$},~~
Ge Zheng\textsuperscript{1},~~
Jiepeng Wang\textsuperscript{3},~~
Zhiyang Dou\textsuperscript{3},~~
Sibei Yang\textsuperscript{1}\thanks{Corresponding author.}\\
\textsuperscript{1}ShanghaiTech University,~~
\textsuperscript{2}The Hong Kong University of Science and Technology,~~\\
\textsuperscript{3}The University of Hong Kong~~
\\
\texttt{\{huanghzh2022, zhengge2023, yangsb\}@shanghaitech.edu.cn},~~\\
\texttt{yuanly@ust.hk},~~\texttt{\{jiepeng, zhiyang0\}@connect.hku.hk}
}

\newcommand{\methodname}{MVTokenFlow\ }

\iclrfinalcopy 
\begin{document}

\maketitle

\begin{abstract}
In this paper, we present \methodname for high-quality 4D content creation from monocular videos. Recent advancements in generative models such as video diffusion models and multiview diffusion models enable us to create videos or 3D models. However, extending these generative models for dynamic 4D content creation is still a challenging task that requires the generated content to be consistent spatially and temporally. To address this challenge, \methodname utilizes the multiview diffusion model to generate multiview images on different timesteps, which attains spatial consistency across different viewpoints and allows us to reconstruct a reasonable coarse 4D field. Then, \methodname further regenerates all the multiview images using the rendered 2D flows as guidance. The 2D flows effectively associate pixels from different timesteps and improve the temporal consistency by reusing tokens in the regeneration process. Finally, the regenerated images are spatiotemporally consistent and utilized to refine the coarse 4D field to get a high-quality 4D field. Experiments demonstrate the effectiveness of our design and show significantly improved quality than baseline methods. Project page: \href{https://soolab.github.io/MVTokenFlow}{https://soolab.github.io/MVTokenFlow}.
\end{abstract}

\section{Introduction}

With the development of generative artificial intelligence (GenAI) technologies, 2D or 3D content creation~\citep{rombach2022latentdiff} already witnessed a huge improvement in recent years. Automatic creation of 4D content~\cite{singer2023text} is an emerging research topic in recent years, which has wide applications in various fields such as AR/VR, video generation, and robotics. However, due to the scarcity of 4D datasets, automatically creating 4D content is still a challenging task.

Due to the huge success of 2D diffusion models for image or video generation, most works~\citep{singer2023text,zhao2023animate124,bahmani20244dfy,zheng2024dreamin4d} focus on how to utilize these 2D diffusion models for 4D content creation. These works aim to generate synchronized multiview videos for a 3D object and then reconstruct a dynamic 3D representation such as 3DGS or NeRF from the multiview videos. However, such a pipeline mainly faces the challenge of maintaining spatial consistency, that multiview videos are spatially consistent on the same timestep, and temporal consistency, that each video is consistent across different frames on different timesteps. Early stage works~\citep{singer2023text,zhao2023animate124,bahmani20244dfy,zheng2024dreamin4d,ling2024alignyourgauss} attain both consistencies by utilizing both spatial and temporal Score Distillation Sampling (SDS) losses. They utilize SDS to distill an image diffusion model~\citep{liu2023zero123,rombach2022latentdiff} to create a 3D representation and then animate the 3D representations by distilling a video diffusion model~\citep{blattmann2023stable}. Based on these SDS pipelines, some works~\citep{jiang2023consistent4d,yin20234dgen,pan2024fastdy4d} explicitly generate a video as guidance and design new 4D representations to learn better motion fields. However, 4D contents created by these SDS-based methods suffer from low-quality motions and over-saturated appearances. Some very recent works~\citep{zhang20244diffusion,liang2024diffusion4d,li2024vividzoo,ren2024l4gm,jiang2024animate3d} directly train a temporally consistent multiview diffusion model from 4D data and reconstruct the 4D representation from the generated multiview images. Though these methods achieve high-quality 4D content, they often require a large amount of 4D training data and extensive computation resources for training.

\begin{figure}
    \centering
    \includegraphics[width=1.0\linewidth]{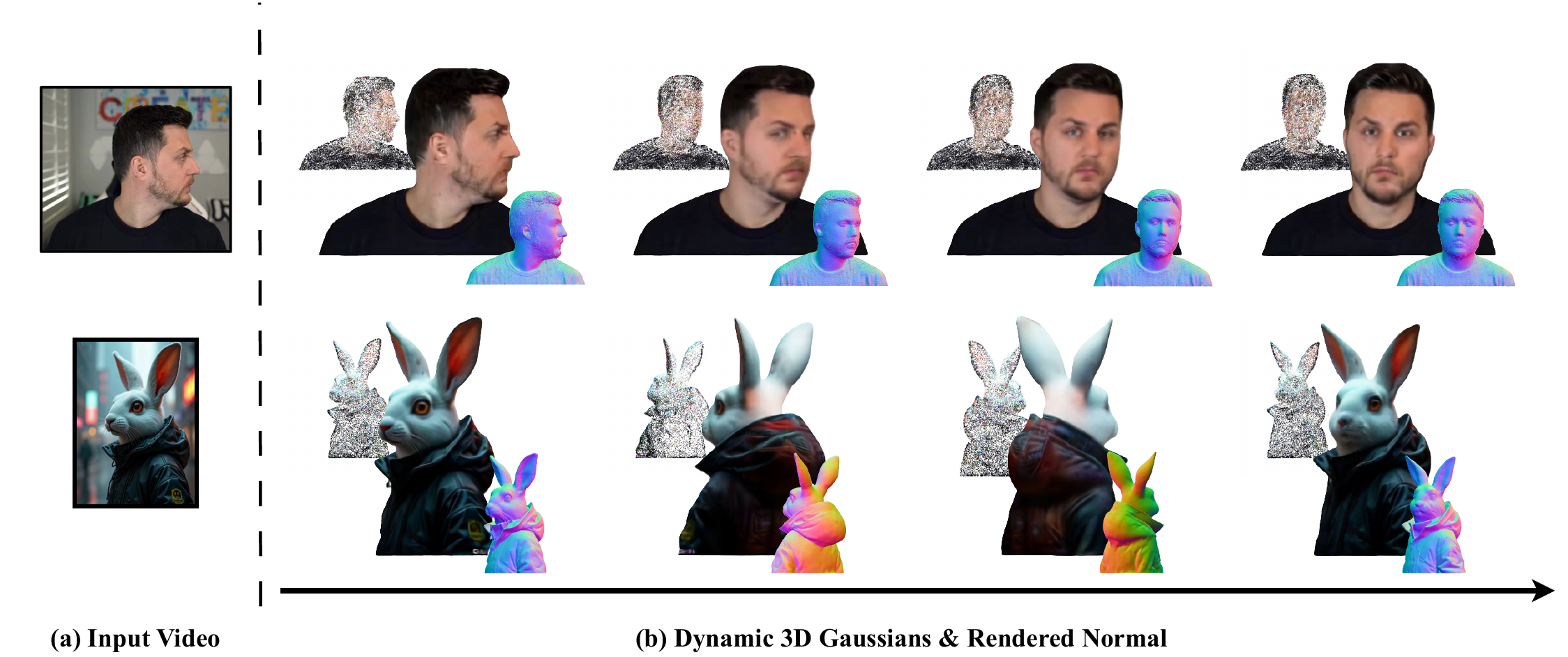}
    \caption{Given an input monocular video containing a foreground dynamic object (left), \methodname generates a 4D video represented by a dynamic 3D Gaussian field (right) by utilizing a multiview diffusion model and a token propagation method to improve both the spatial and temporal consistency. On the right, we also show the colors of these Gaussian spheres and the rendered normal maps besides the rendered RGB images.}
    \label{fig:teaser}
\end{figure}

In this paper, we propose a novel 4D generation framework called \methodname based on multiview diffusion models~\citep{li2024era3d} and the token flow technique~\citep{geyer2023tokenflow}. 
The challenge in high-quality 4D generation is to create temporally consistent multiview images from monocular 2D videos. Our core motivation is to first construct a coarse 4D field, temporally consistent in the front view, and refine it to achieve high temporal consistency across all views. Specifically, in the coarse stage, we utilize multiview diffusion and the temporal consistency of the front view to enhance the spatiotemporally consistent 4D field for better rendered 2D flow. In the refinement stage, this rendered 2D flow guides the regeneration of consistent multiview images, culminating in a high-quality 4D field.
As shown in Fig.~\ref{fig:teaser}, given an input video that can be generated from a text- or image-to-video generative model, \methodname generates high-quality 4D contents represented by a dynamic Gaussian field that can be rendered with the splatting technique on arbitrary viewpoints and timesteps. \methodname achieves spatial consistency by applying the multiview diffusion model to generate multiview images on every frame. The multiview diffusion model is trained on a large 3D dataset to preserve the spatial consistency among all the multiview images. Then, these multiview images will be used in the training of an initial dynamic Gaussian field as the coarse 4D field.

Since the multiview diffusion model is applied to each frame separately, the generated multiview images on different time steps are not temporal consistent with each other, which causes blurry renderings of the coarse 4D field. To improve the temporal consistency between different timesteps, we further apply token flow to associate the generated images of the same viewpoint but on different timesteps. This is based on the fact that a temporally consistent video should share similar tokens among different frames, which has already been utilized by the Token Merging~\citep{li2024vidtome} or Token Reduction~\citep{geyer2023tokenflow} for video editing. To determine the similar tokens on different frames, we utilize the rendered 2D flows from the coarse 4D field to associate pixels from different frames. Then, we regenerate all the multiview images on all timesteps and force the associated pixels to have similar tokens in the reverse diffusion sampling process. This token flow technique effectively improves the temporal consistency of the regenerated multiview images. Finally, the coarse 4D field is refined by the regenerated images to get a high-quality 4D field.

Our contributions can be summarized as follows: (1) We extend the 2D token flow to multiview diffusion to improve temporal consistency. (2) We design a novel pipeline that interleaves dynamic Gaussian field reconstruction with Multiview Token flow to generate multiview consistent videos. (3) We have shown improved quality in 4D content creation from a monocular video.
We conduct experiments on both the Consistent4D~\citep{jiang2023consistent4d} dataset and a self-collected dataset to validate the effectiveness of our methods. The results demonstrate that our method generates videos with high-fidelity and high-quality motion on unseen views.
\section{Related work}

In recent years, diffusion models have gained prominence as a powerful generative framework, excelling in tasks such as image~\citep{rombach2022latentdiff,nichol2021improvedddpm,nichol2021glide,ramesh2022hierarchicaldiff} and video synthesis~\citep{blattmann2023alignlatentvd,an2023latentshiftvd,ge2023preservecovd,guo2023animatediffvd,singer2022makeavideovd}. These models generate data by progressively denoising randomly initialized samples until a coherent structure or scene emerges. Leveraging the flexibility and effectiveness of generative models, they have been adapted to a wide range of tasks~\cite{zheng2023ddcot, tang2023cotdet, shi2024part2object, tang2023temporal, tang2023contrastive}, including 3D and 4D content generation. 
In this section, we will review three parts: diffusion models, 4D scene representations, and 4D generation with diffusion models.

\paragraph{Diffusion for Generation}
Recently, diffusion models, pre-trained on large-scale datasets~\citep{schuhmann2022laion}, have made significant strides in generating high-quality and diverse visual content for both 2D image and video tasks ~\citep{rombach2022latentdiff,nichol2021improvedddpm,blattmann2023alignlatentvd,an2023latentshiftvd,huang2024free}. Leveraging aligned vision-language representations~\cite{shi2024plain, dai2024curriculum, shi2024devil, shi2023logoprompt, shi2023edadet, shi2022spatial}, these models can produce various forms of visual content with impressive diversity and realism conditioned on text or images. To adapt 2D diffusion models for 3D generation, some methods utilize Score Distillation Sampling Loss~\citep{poole2022dreamfusion,lin2023magic3d,chen2023fantasia3d,wang2024prolificdreamer} to distill 3D priors and train a neural radiance field~\citep{mildenhall2020nerf} for 3D asset creation. However, this approach often faces challenges such as slow training speeds and multi-face artifacts~\citep{shi2023mvdream}. To address these limitations, another strategy involves fine-tuning pre-trained 2D diffusion models to directly generate multi-view consistent images~\citep{shi2023mvdream,liu2023zero123,long2024wonder3d,liu2023syncdreamer,li2024era3d} from large-scale multi-view datasets~\citep{deitke2023objaverse}. These images are then processed through 3D reconstruction algorithms~\citep{wang2021neus,kerbl20233dgs,liu2023nero} to produce high-quality 3D assets. Despite these advancements, efficiently leveraging these techniques for 4D generation, ensuring both spatial and temporal coherence, remains a challenging problem.

\paragraph{4D Scene Representation}
Current 4D scene representations can be broadly categorized into two types based on their underlying 3D scene representation: 1) NeRF-based \citep{mildenhall2020nerf} and 2) 3D Gaussian Splatting (3DGS)-based \citep{kerbl20233dgs}. Both approaches extend static 3D scene representations into the temporal domain by introducing deformable fields or animation-driven training frameworks.
NeRF (Neural Radiance Fields) was initially proposed to encode the geometry and appearance of static scenes using implicit models with MLPs. Building upon this, many works have extended static NeRF to handle dynamic scenes, either by modeling a dynamic deformation field on the top of a canonical static scene representation~\citep{pons2021dnerf,tretschk2021nonrigidnerf,yuan2021stardnerf,park2021nerfies,fang2022fastdnerf} or by directly learning a time-conditioned radiance field~\citep{li2022neural3dvideo,gao2021dynamicviewsynthesis,park2021hypernerf,xian2021spacenerfvideo}. Despite its success, NeRF-based methods often face limitations in training and inference speed, making them less suitable for real-time applications.
Recently, 3D Gaussian Splatting (3DGS) has shown impressive performance due to its efficient training and real-time novel view synthesis capabilities. This method represents static scenes as a set of Gaussian primitives and employs a fast Gaussian differentiable rasterizer with adaptive density control. As an explicit representation, 3DGS also simplifies tasks such as scene editing. 
3DGS then has been applied to model dynamic scenes with the similar idea of building a deformation field~\citep{luiten2023dynamic3dgauss,wu20244dgauss,yang2024deformable4dgauss,zeng2024stag4d,wu2024sc4d}.
For example, Dynamic 3D Gaussians~\citep{luiten2023dynamic3dgauss} enable the Gaussians to move and rotate over time under local rigid constraints. This approach efficiently models fine details and temporal changes, making it highly effective for 4D content creation.
Together, these representations offer a robust framework for generating realistic and temporally coherent dynamic scenes in 4D space, supporting applications such as animation, scene reconstruction, and motion capture.

\paragraph{4D Generation} By efficiently integrating advanced diffusion techniques with 4D scene representations, significant progress has been made toward 4D generation. One approach in this direction leverages Score Distillation Sampling~\citep{poole2022dreamfusion} to distill spatial and temporal prior knowledge from multiple diffusion models into a 4D scene representation, producing spatially and temporally consistent 4D objects, including text-to-video and text-to-image generation. 
A pioneering work, MAV4D~\citep{singer2023mav3d} introduced a multi-stage training pipeline for dynamic scene generation, utilizing a Text-to-Image (T2I) model to initialize static scenes and a Text-to-Video (T2V)~\citep{singer2022makeavideovd} model to handle motion dynamics.
Building on this paradigm, several methods have sought to improve 4D generation quality by incorporating image conditions ~\citep{zhao2023animate124}, hybrid Score Distillation Sampling~\citep{bahmani20244dfy}, strategies that decouple static elements from dynamic ones~\citep{zheng2024dreamin4d}, and related techniques. However, these methods are largely based on NeRF variants, which suffer from issues like over-saturated appearance and long optimization times. To overcome these limitations,  Align-Your-Gaussians~\citep{ling2024alignyourgauss} proposed using dynamic 3D Gaussian Splatting (3DGS)~\citep{kerbl20233dgs} as the underlying 4D scene representation to learn a deformation field~\citep{park2021nerfies,pons2021dnerf}, offering faster training and better real-time capabilities. Despite this, the reliance on SDS loss in these methods leads to slow optimization speeds, limiting their applicability in downstream tasks.
Another approach uses video as guidance. Several video-to-4D frameworks~\citep{jiang2023consistent4d,yin20234dgen,pan2024fastdy4d} have been introduced that use video inputs as references to guide 4D generation. These methods attempt to generate dynamic scenes by leveraging video-driven information for more precise motion dynamics.
Additionally, to ensure multi-view consistency, recent works have focused on retraining multi-view video diffusion models~\citep{zhang20244diffusion,liang2024diffusion4d,li2024vividzoo,ren2024l4gm,jiang2024animate3d} with 4D datasets, integrating both spatial and temporal modules. However, these models often require large amounts of data and are computationally intensive.


\section{Method}

\begin{figure}
    \centering
    \includegraphics[width=0.98\linewidth]{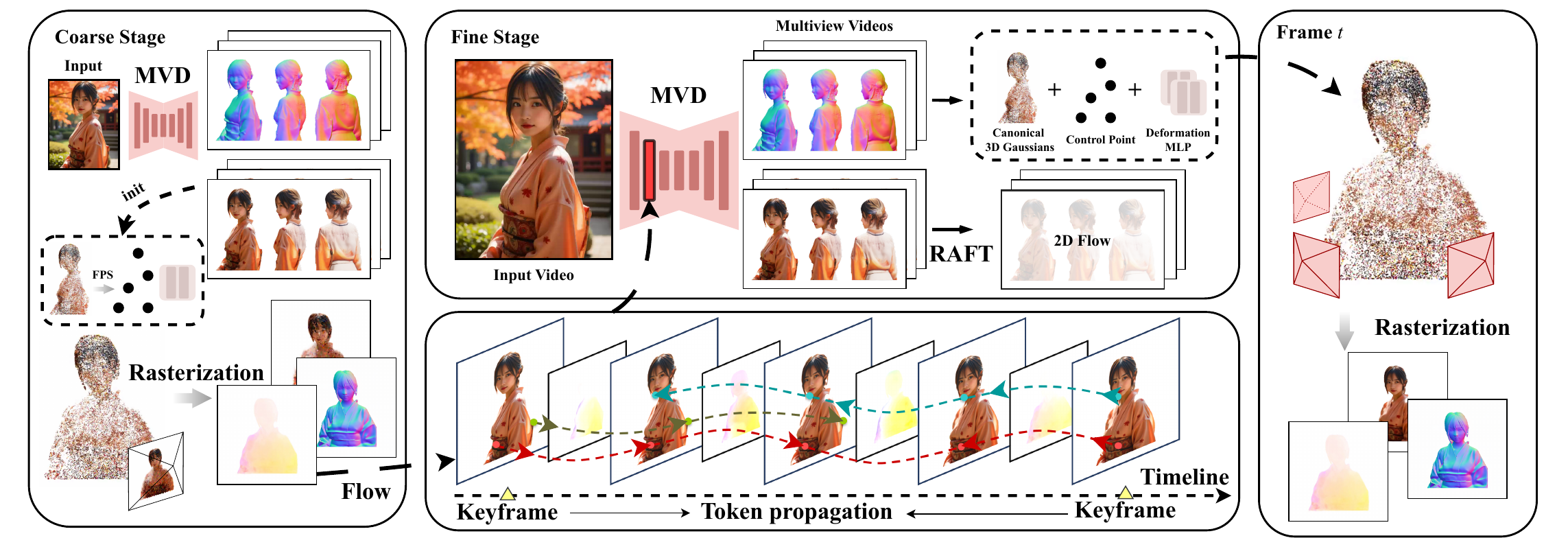}
    \caption{\textbf{Overview}. Given an input video that can be generated by video diffusion models, we first apply the Era3D~\cite{li2024era3d} to generate the multiview-consistent images and normal maps for each timestep. Then, we reconstruct a coarse dynamic 3D Gaussian field field from the generated multiview images. After that, we use the coarse dynamic 3D Gaussian field to render 2D flows to guide the re-generation of the multiview images of Era3D, which greatly improves the temporal consistency and image quality. Finally, the regenerated images are used in the refinement of our dynamic 3D Gaussian field to improve the quality.}
    \label{fig:overview}
\end{figure}

Given a monocular video with a dynamic foreground object, our target is to reconstruct the dynamic 3D field represented by a static 3D Gaussian field (3DGS)~\citep{kerbl20233dgs} and a time-dependent deformation field to deform the 3D Gaussian field to a specific timestep. Note that the video can be either provided or generated from a text description or an image by video diffusion models.
Fig.~\ref{fig:overview} illustrates the outline of our pipeline with 3 steps.
First, we run a pretrained multiview diffusion model Era3D~\citep{li2024era3d} to generate multiview videos on predefined viewpoints (Sec.~\ref{sec:mv-init}). 
Then, we reconstruct a coarse dynamic 3D field from the generated multiview videos in Sec.~\ref{sec:recon}, where we introduce the representation and training process.
Finally, we regenerate the multiview video by combining the pretrained multiview diffusion model with the token flow with the rendered 2D flow from our coarse dynamic 3D field (Sec.~\ref{sec:mv-re}). The regenerated images are used in the refinement of our coarse 4D field to an improved 4D field with better quality and consistency.

\subsection{Multiview Video Generation}
\label{sec:mv-init}

In this stage, we introduce the pretrained multiview diffusion model Era3D~\citep{li2024era3d} to generate multi-view videos from the input monocular video, which will be used in the supervision of dynamic 3D Gaussian reconstruction. 

\textbf{Multiview diffusion model}. Given a reference video, we split it into $N$ frames $I_{\text{ref}}^{(n)}\in \mathcal{R}^{H\times W\times 3}$ with $n=1,...,T$. For every frame, we apply the multiview diffusion model Era3D~\citep{li2024era3d} to generate multiple novel view images $I^{(n,k)}$ where $k=1,...,K$ is an index of the viewpoint. Note that combining all the generated frames on the same viewpoint but different time steps leads to $K$ videos on different viewpoints. Era3D not only estimates the RGB images but also predicts normal maps on these viewpoints. We use the same symbols to denote both the normal maps and RGB images.

\textbf{Discussion about temporal inconsistency}. Simply applying the multiview diffusion model to every frame maintains the consistency between all images of different viewpoints $I^{(n,1:K)}:=\{I^{(n,k)}|k=1,...,K\}$ but loses coherence between images $I^{(n_0,k)}$ and $I^{(n_1,k)}$ from two arbitrary different timesteps $n_0$ and $n_1$. The reason is that the multiview diffusion model can only maintain the consistency between different viewpoints but the independent generation on different timesteps leads to temporal inconsistency. Thus, the key problem is to improve the temporal consistency of the generated videos. An important observation is that for two frames of a plausible video, their diffusion features or so-called tokens share a strong similarity and are correlated by the 2D flow between them~\citep{geyer2023tokenflow}. This attribute has been observed by many video editing papers~\citep{geyer2023tokenflow,li2024vidtome} to design token merging~\citep{li2024vidtome} and token flow~\citep{geyer2023tokenflow}. Based on this observation, we improve the temporal consistency by the following two strategies, i.e. enlarged self-attention layers and token propagation with 2D flows.

\textbf{Enlarged self-attention}. As observed by many previous works, enlarging the self-attention of stable diffusion~\citep{rombach2022latentdiff} (SD) to all the images of different timesteps is helpful in improving the temporal consistency. The adopted Era3D model is also based on the SD model so we enlarge the self-attention layers to include all timesteps to improve temporal consistency. Specifically, in each self-attention layer of the image $I^{(n,k)}$, we keep the query features unchanged but adopt all the features from images $I^{(m,k)}$ of the same $k$-th viewpoint but different timesteps $m$ as the keys and values. This enlarged self-attention provides free temporal consistency without retraining the diffusion model.

\textbf{Token propagation with 2D flows}.
To further improve the consistency, for a specific video on a specific viewpoint, we only conduct denoising on several keyframes and then propagate the features (tokens) of keyframes to the rest frames.
Although minor inconsistencies may exist in the keyframes, we can still reconstruct a high-quality 4D field because keyframes are derived from a temporally consistent input video, and the dynamic 3D Gaussian field is supervised by the video, smoothing residual inconsistencies.
Specifically, to conduct denoising on the video $I^{(1:N,k)}_t$ corresponding to the $k$-th viewpoint to get $I^{(1:N,k)}_{t-1}$, we first sample $M$ equidistant keyframes $\{I^{(n_m,k)}_t|m=1,2,...,M\}$ and we conduct the normal denoising process on all these keyframes to get their denosied $\{I^{(n_m,k)}_{t-1}\}$. We obtain the self-attention features $F^{(n_m,k)}_t$ of all these keyframes. Then, for the rest frames, we propagate the features of keyframes to denoise them. Specifically, for a specific frame $I_t^{(n,k)}$ with $n_{m-1}\le n \le n_{m}$, we utilize the 2D flows $\pi({n_{m-1}\to n})$ and $\pi(n \to n_{m})$ to warp the features, resulting warped features $F^{(n_{m-1}\to n,k)}_t$ and $F^{(n_{m}\to n , k)}_t$. We compute the features on the  $I_t^{(n,k)}$ by
\begin{equation}
    F^{(n,k)}_t =(1-\lambda_{n}) \cdot F^{(n_{m-1}\to n,k)}_t + \lambda_{n} F^{(n_{m}\to n,k)}_t,
    \label{eq:propagate}
\end{equation}

where $\lambda_n = (n_{m}-n)/(n_{m}-n_{m-1})$ is a position-dependent weighting parameter. Then, we use these propagated features to denoise these intermediate frames between keyframes. Due to the presence of regions in the video that cannot be covered by optical flow, we only propagate features in the early stages $t\le \tau$, allowing the diffusion process to add more details and occluded regions. This token propagation scheme effectively utilizes the redundancy in a video and has the potential to improve the temporal consistency of the generated video. Implementing the token propagation requires the estimation of the 2D flow, which is introduced in the following.

\textbf{Estimation of 2D flows}. In the beginning, we only have access to the input video but do not have any information on other unseen viewpoints. Thus, we estimate the 2D flow of the input reference video by RAFT~\citep{teed2020raft}. Then, we use this estimated 2D flow to propagate the features of the video on the first viewpoint, i.e. the front view of Era3D, while for other viewpoints, we apply the full denoising process for all diffusion timesteps. Though only one 2D flow is utilized on the front view, the Era3D will utilize cross-viewpoint attention layers to propagate the consistency of the front view to other views and thus improve the temporal consistency.

\subsection{Reconstruction with Gaussian Splatting}
\label{sec:recon}

Given the generated multiview videos, in this stage, we aim to reconstruct a dynamic 3D Gaussian field. Our method employs a keypoint-controlled dynamic 3D Gaussian representation~\citep{huang2024sc}, comprising a static 3D Gaussian and a time-dependent deformable field. For the static 3D field, we represent the field as a set of 3D Gaussians as proposed in 3DGS~\citep{kerbl20233dgs}. For the deformation field, we adopt the representation from SC-GS~\citep{huang2024sc}, which first generates a set of 3D control points by clustering 3D Gaussians, then applies an MLP network to translate and rotate these control points, and finally deforms the 3D Gaussians with these control points. On each control point, we associate a set of learnable radius parameters of a radial-basis-function (RBF) kernel that controls how the impact of the control point on a
Gaussian will decrease as their distances increase. We train this dynamic Gaussian field using the generated multiview videos. Besides the rendering loss, mask loss, structural dissimilarity (D-SSIM)~\citep{kerbl20233dgs} loss, and as-rigid-as-possible (ARAP) loss, we also adopt a normal map loss and a 2D flow loss.

\textbf{Flow loss}. The flow loss here is to minimize the difference between the rendered 2D flows and the estimated 2D flows on the front view by RAFT~\citep{teed2020raft}. Specifically, for two timesteps, we project the 3D offset of each 3D Gaussian onto to image plane to get the 2D offset of the 3D Gaussian. Then, we combine these 2D offsets with the same alpha blending method as used in splatting to render a 2D flow map. We minimize the difference between the rendered 2D flow map and the estimated 2D flow map with an L1 loss and skip the invisible regions caused by occlusions. 

\textbf{Normal loss}. Since the Era3D model also generates normal maps for every viewpoint, we also supervise the dynamic 3D Gaussian field with these normal maps. However, computing normal maps from 3DGS is ambiguous without a clear definition of the normal directions. Instead, we first render the depth maps by alpha blending the depth values of all 3D Gaussians and then compute a normal map from the rendered depth map. Finally, we minimize the difference between the generated normal maps and the rendered normal maps. These normal maps pose a geometric constraint on the dynamic 3D Gaussian field and improve the rendering quality.

\subsection{Regeneration with 2D flows}
\label{sec:mv-re}
In this section, we regenerate the multiview images by Era3D with the help of the coarse dynamic 3D Gaussian field trained in the previous section. Then, these regenerated images are used in the refinement of the coarse dynamic 3D Gaussian field. The coarse dynamic 3D Gaussian field often produces blurry results because the images generated in Sec.~\ref{sec:mv-init} are not temporally consistent enough. However, we observe that these coarse 3D Gaussian fields already produce a reasonable 3D flow field that could improve the temporal consistency for the multiview generation. Thus, we regenerate all the multiview videos with the help of the 3D flow field and refine the dynamic 3D Gaussian field with the regenerations.

\textbf{Regeneration and refinement}. In Sec.~\ref{sec:mv-init}, we only have access to the 2D flow of the front view to guide the generation of the multiview videos so all these unseen views are not well constrained with temporal consistency. With the coarse dynamic 3D Gaussian field, we render 2D flow maps for all viewpoints and then incorporate the token propagation in the generation process as introduced in Eq.~(\ref{eq:propagate}). By propagating tokens on all viewpoints, we greatly improve the temporal consistency of all generated videos. Then, we utilize the regenerated videos to refine our dynamic 3D Gaussian field, which achieves better rendering quality with less blurry results.

\section{Experiment}

\begin{figure}[t]
    \centering
    \includegraphics[width=0.98\linewidth]{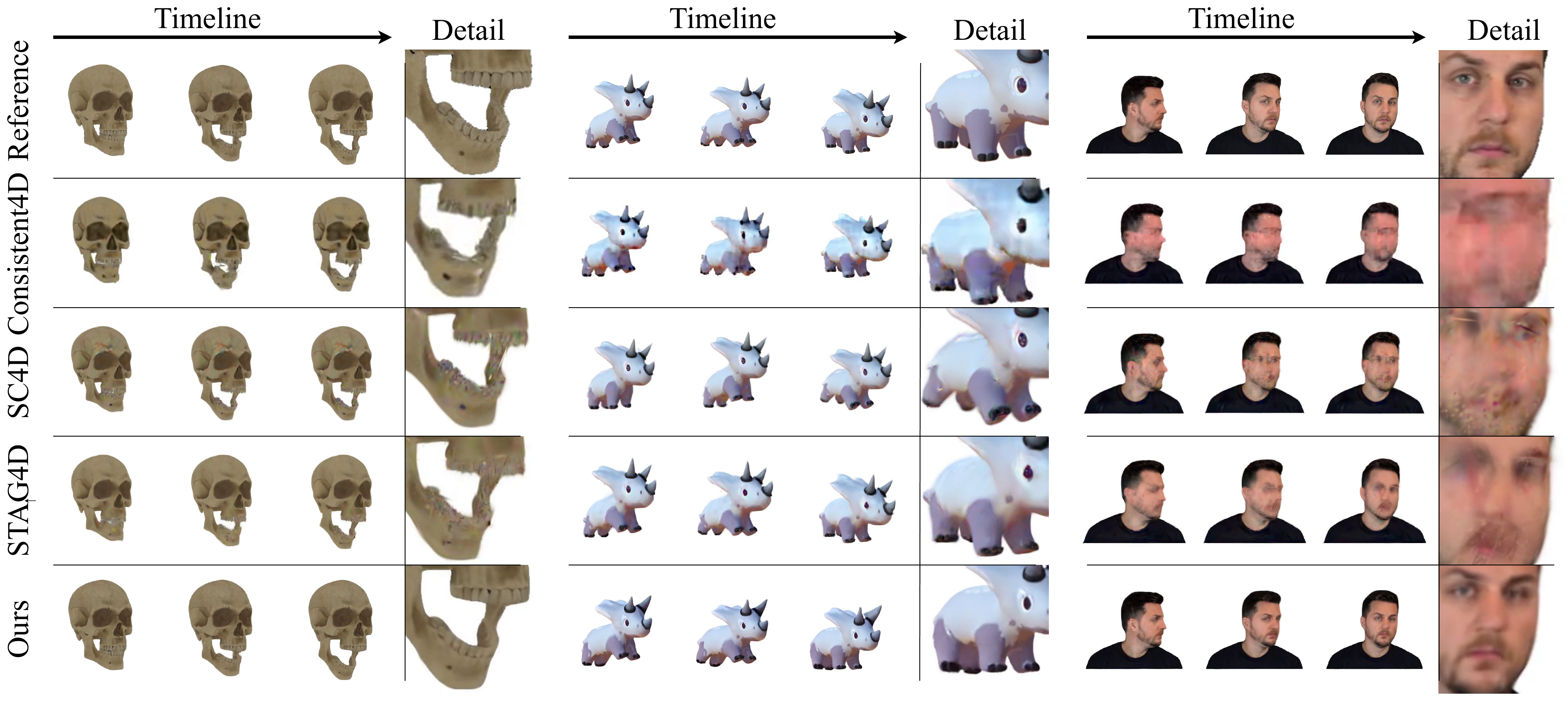}
    \caption{Qualitative comparison on temporal consistency of our method with baseline methods, Consistent4D~\citep{jiang2023consistent4d}, SC4D~\citep{wu2024sc4d}, and STAG4D~\citep{zeng2024stag4d}.}
    \label{fig:exp-time}
\end{figure}

\begin{figure}[t]
    \centering
    \includegraphics[width=0.98\linewidth]{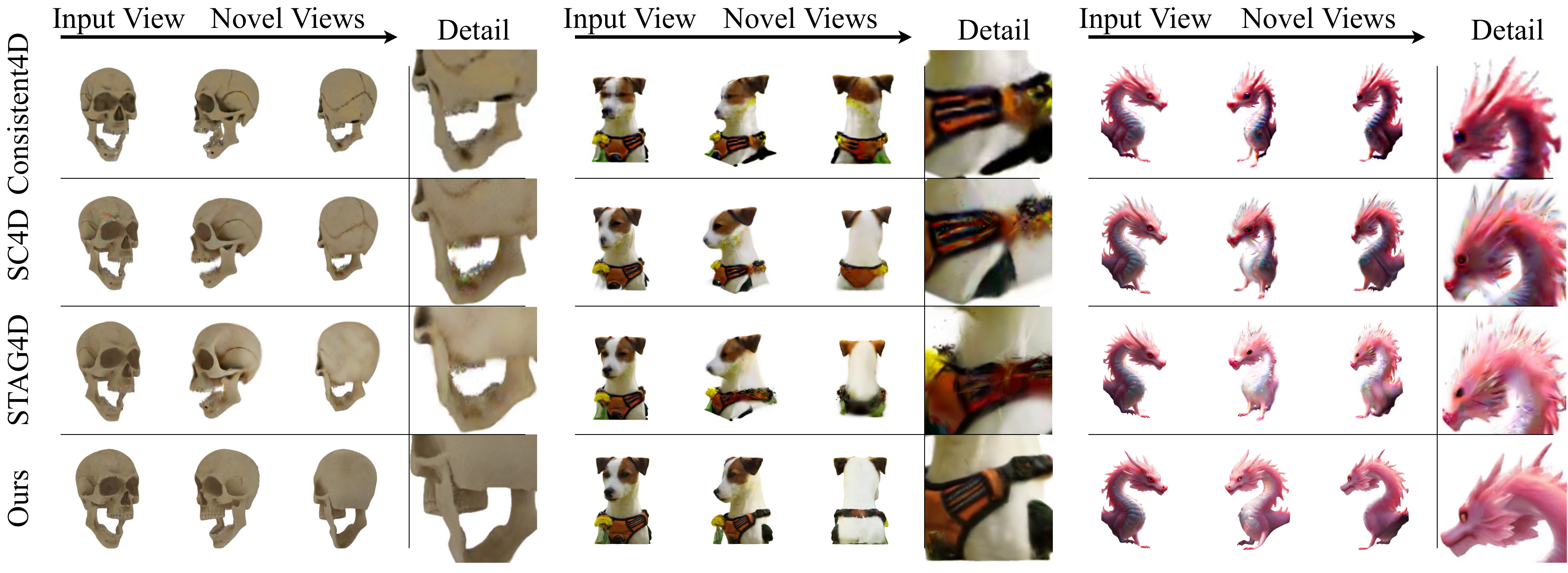}
    \caption{Qualitative comparison on spatial consistency of our method with baseline methods, Consistent4D~\citep{jiang2023consistent4d}, SC4D~\citep{wu2024sc4d}, and STAG4D~\citep{zeng2024stag4d}.}
    \label{fig:exp-view}
\end{figure}
\subsection{Experimental Settings}

\textbf{Implementation details.} For multi-view video generation, we utilize Era3D~\citep{li2024era3d} to generate $K=6$ viewpoints at a resolution of 512x512 for one frame, using 40 denoising steps. We set $\tau =20$, executing token propagation during the denoising process when $t<\tau$. For keyframe selection, we employ a keyframe interval of 8 frames. In the initialization phase of dynamic 3D Gaussian representation, we initialize 512 control points with Farthest Point Sampling (FPS) sampling. Each Gaussian point is influenced by its 3 nearest control points. During the training of the dynamic Gaussian field, we use an initial learning rate of $3\times 10^{-4}$ for the MLP, followed by exponential decay. In the refinement phase of the dynamic 3D Gassuian field with regenerated multiview videos, we reset the learning rate to its initial value and applied the same decay strategy. All experiments are conducted on an NVIDIA A40 GPU. We provide more implementation details in the appendix about the losses and data preparation.

\textbf{Dataset.} For quantitative comparisons and part of qualitative comparisons, we evaluate our method with the dataset from Consistent4D~\citep{jiang2023consistent4d}. The dataset comprises 12 synthetic videos and 12 in-the-wild videos. All videos are monocular, captured by a stationary camera, consisting of 32 frames and lasting approximately 2 seconds. We also collect some videos on the Internet for evaluation.

\textbf{Evaluation metrics.} We evaluate our method from three perspectives: consistency with reference videos, spatial consistency, and temporal consistency. Following Consistent4D~\citep{jiang2023consistent4d}, we utilize PSNR, SSIM, and LPIPS to assess the consistency with reference videos. For multi-view consistency, we employ the CLIP score to measure the semantic similarity of images from different viewpoints.

\begin{table*}[t]
\caption{Quantitative results across the three evaluation perspectives, view synthesis, spatial consistency and temporal consistency.}
\setlength{\tabcolsep}{0.25em} %
\centering
\begin{tabular}{l|ccc|c}
\toprule
\multirow{2}{*}{\textbf{Method}} & \multicolumn{3}{c|}{\textbf{View Synthesis}} & \multicolumn{1}{c}{\textbf{Spatial Consisency}} \\
 & LPIPS $\downarrow$ & PSNR $\uparrow$ & SSIM $\uparrow$   & CLIP $\uparrow$  \\
\midrule
Consistent4D~\citep{jiang2023consistent4d} & 0.09 & 23.97 & 0.91 & 0.89  \\
SC4D~\citep{wu2024sc4d} & 0.08 & 29.50 & 0.95 & 0.90  \\
STAG4D~\citep{zeng2024stag4d}  & 0.06 & 30.79 & 0.94 & 0.91 \\
Ours & \textbf{0.04} & \textbf{31.27} & \textbf{0.97} & \textbf{0.91}  \\
\bottomrule
\end{tabular}%
\label{tab:quantity}
\end{table*}

\subsection{Comparisons} We compare our \methodname with recent available open-source methods, including Consistent4D~\cite{jiang2023consistent4d}, SC4D~\cite{wu2024sc4d} and STAG4D~\cite{zeng2024stag4d}. Experiments are conducted on the Consistent4D dataset and a selection of collected videos. This section analyzes representative qualitative results and quantitative results, with additional visualization available in the supplementary videos. We also provide more results of our method in Sec.~\ref{sec:addtional-results} of appendix.

\begin{figure}[b]
    \centering
    \includegraphics[width=0.98\linewidth]{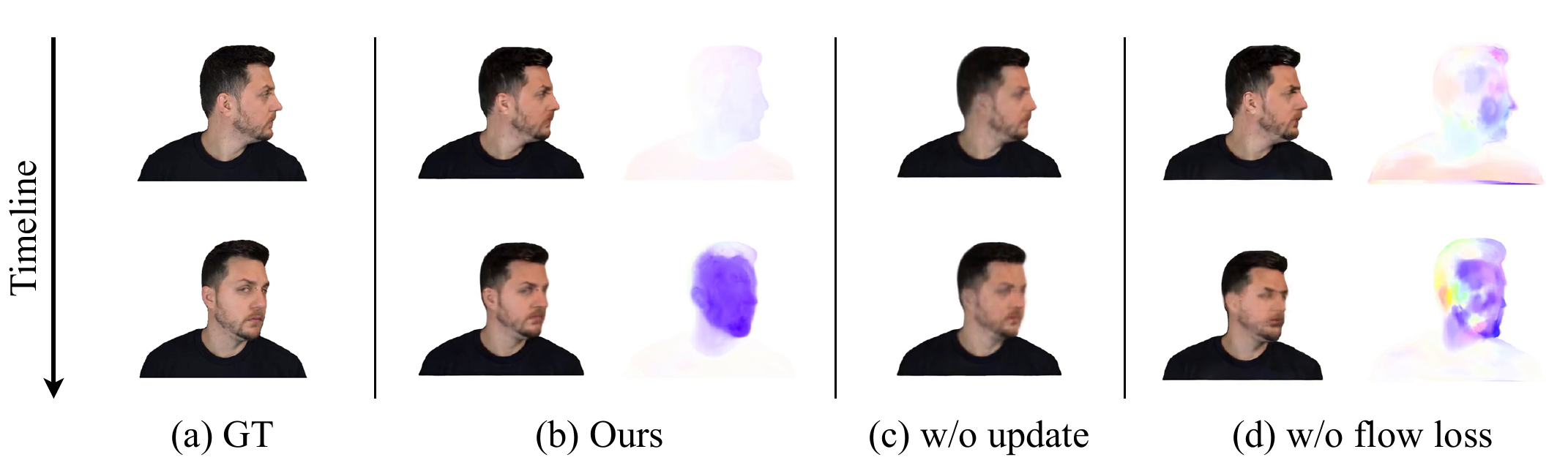}
    \caption{Ablation study of the overall architecture. The four parts illustrate (a) Input viewpoint. (b) Our final results. (c) The intermediate outcome from our coarse dynamic 3D field. (d) Result without flow loss.}
    \label{fig:ablation}
\end{figure}
\textbf{Qualitative Comparison on Temporal Consistency.} Fig.~\ref{fig:exp-time} shows the comparison from the perspective of temporal consistency on three samples (\textit{skull, triceratops, and man turning his head}). Consistent4D~\cite{jiang2023consistent4d} is limited by the expressiveness of cascaded DyNeRF~\cite{li2022neural3dvideo}, making it challenging to represent the textures of dynamic objects. As a result, the generated outputs are often blurry and distorted, accompanied by artifacts and color discrepancies. SC4D struggles to model motion; as demonstrated in the example of the \textit{triceratops}, its results for the second and third frames are inconsistent with the reference viewpoint due to the weak temporal consistency by simply distilling video diffusion models. For complex real-world scenarios, such as the example of \textit{man turning his head}, early methods are unable to generate coherent and realistic motions. Though recent work (STAG4D) achieves satisfactory temporal consistency in simple samples, it suffers from significant blurriness and distortion when modeling the man turning his head. This failure may be attributed to their reliance on the temporal consistency provided by the single-view reference video, which hinders their ability to learn the correct 3D temporal variations.

Compared to previous methods, we adopt flow loss in the generation process, decoupling the learning of motion and appearance. Such a strategy allows our \methodname to capture high-quality motions while simultaneously modeling clear and detailed appearances.

\textbf{Qualitative Comparison on Spatial Consistency.} We also perform comparisons with other methods across different viewpoints to demonstrate spatial consistency of our method, as shown in Fig.~\ref{fig:exp-view} with three samples (\textit{skull, dog and dragon}). 
For novel viewpoints generation, both SC4D and STAG4D tend to produce artifacts, resulting in a significant performance drop compared to the input view. In the example of the \textit{dog}, other methods struggle to maintain the pattern of the bag the dog is wearing, often generating blurred or noisy results. This phenomenon becomes even more pronounced with complex inputs (\textit{dragon}), where previous methods fail to generate eyes that correspond with the input view and similar whiskers. Meanwhile, the generations show over-saturated colors and some artifacts with inconsistent colors.

\begin{figure}[t]
\begin{floatrow}

\capbtabbox{

\begin{tabular}{lccc}
\toprule
 & LPIPS $\downarrow$ & CLIP $\uparrow$ & FVD $\downarrow$    \\
\midrule
DG4D & 0.1748 & 0.915 & 856.86  \\
Consistent4D & 0.1729 & 0.865 & 1072.94  \\
SC4D & 0.1659 & 0.915 & 879.66 \\
STAG4D & 0.1506 & 0.885 & 972.73  \\
Ours & \textbf{0.1216} &  \textbf{0.948} & \textbf{846.32}  \\
\bottomrule
\end{tabular}%
}{%
  \caption{Quantitative results of novel view synthesis on Consistent4D synthetic objects with multi-view videos.}%
  \label{tab:exp_novel_view}
}

\ffigbox{
\includegraphics[width=0.4\textwidth]{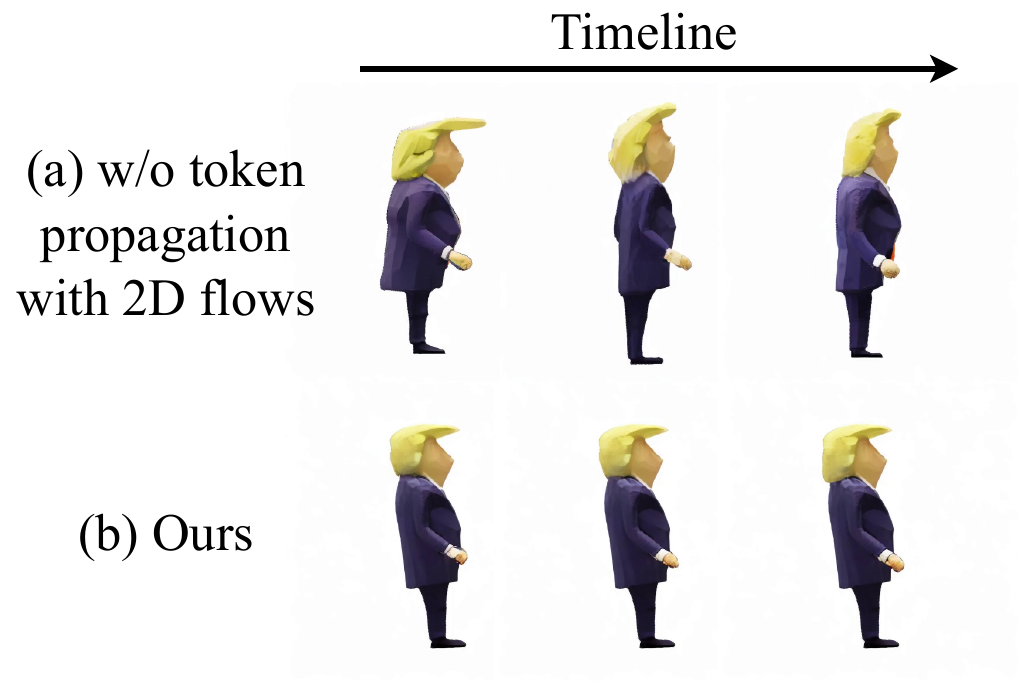}
}{
    \caption{Ablation study on token propagation for multiview video generation.}
    \label{fig:abl_mvd}
}

\end{floatrow}
\end{figure}

In comparison, our \methodname outperforms these methods in both fine-grained spatial consistency and generation quality. These advantages result from our utilization of multiview diffusion combined with normal loss and rendered 2D flow maps for all viewpoints, which leads to strong geometry constraints, whereas previous methods rely on SDS loss as the source of multi-view consistency.

\textbf{Quantitative Comparison.} For quantitative comparisons, we evaluate metrics introduced above on Consistent4D dataset and the results are shown in Table~\ref{tab:quantity}. Across all metrics, our method consistently outperforms previous approaches including Consistent4D, SC4D and STAG4D. As mentioned in the metric introduction, these results demonstrate that our method exhibits superior consistency with the reference video and enhanced spatiotemporal coherence. Notably, the improvement in LPIPS, which reveals the perceptual consistency of generated images, is particularly significant. This demonstrates that our method can produce more realistic results, aligning with the detailed and accurate textures showcased in our qualitative analysis.
Furthermore, as shown in Table~\ref{tab:exp_novel_view}, we evaluated our method in terms of novel view video synthesis result on Consistent4D synthetic dataset. Our approach exhibits superior temporal consistency under novel views when compared to recent methods.

\subsection{Ablation Study}

To demonstrate the effectiveness of token propagation in the multiview video generation stage and the overall architecture we proposed, we conducted ablation studies for each aspect.

First, we ablate token propagation for multiview video generation as shown in Fig.~\ref{fig:abl_mvd}. Without token propagation with 2D flows, both the pose and body shape of the \textit{Trump} exhibit significant variations across different time steps, resulting in a video with flickering changes. In contrast, our proposed method effectively constrains the phenomenon of flickering, significantly improving the temporal consistency of multiview video generation.

\begin{figure}[h]
\begin{floatrow}

\capbtabbox{%
  \begin{tabular}{lccc}
    \toprule
     & LPIPS $\downarrow$ &  FVD $\downarrow$  & CLIP $\uparrow$    \\
    \midrule
    w/o Enlarged SA & 0.0425 & 348.36  & 0.881 \\
    w/o Flow & 0.0408 & 341.50 & 0.887 \\
    w/o Noraml  & \textbf{0.0405} & 320.93 & 0.885  \\
    Full & 0.0423 &  \textbf{313.47} & \textbf{0.891}  \\
    \bottomrule
    \end{tabular}%
}{%
  \caption{Ablation study on different components in the Consistent4D Dataset.}%
    \label{tab:ablation}
}

\ffigbox{
    \includegraphics[width=0.4\textwidth]{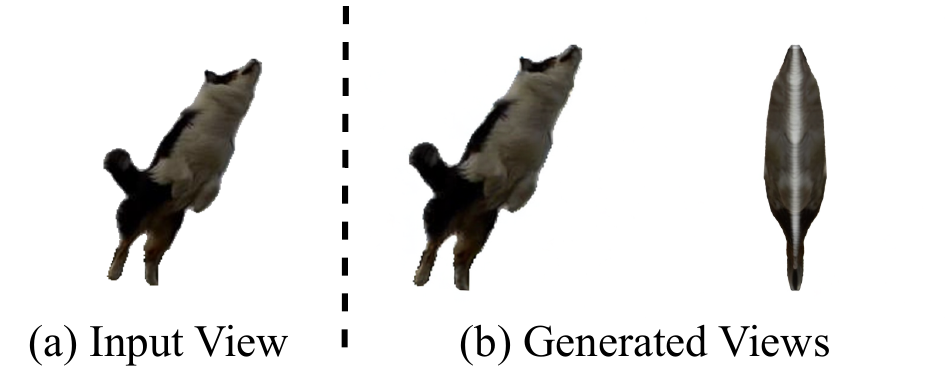}
}{
    \caption{Limitation on generating novel views for uncommon viewpoints.}
    \label{fig:limitation}
}

\end{floatrow}
\end{figure}

Then, we remove the flow loss or skip the regeneration and refinement phase, and compare the results with our full model. To analyze the impact on both motion and appearance, we present both rendered RGB images and extracted optical flow at different timesteps in Fig.~\ref{fig:ablation}. Comparing (b) with (d), the flow loss significantly enhances the quality of the extracted optical flow, indicating more reliable 3D temporal variations. While (c) displays a relatively blurry image compared to the other experiments, highlighting the significant improvement in generation quality during the regeneration and refinement phase.

Table~\ref{tab:ablation} presents the quantitative results of our ablation study on the Consistent4D~\cite{jiang2023consistent4d} dataset across three metrics: reference view alignment (LPIPS), temporal consistency (FVD), and multi-view consistency (CLIP). The results demonstrate that flow propagation improves temporal consistency, while the normal loss contributes to enhancing multi-view consistency.

\vspace{-2mm}
\section{Limitations}
Though \methodname succeeds in reconstructing a 4D video from a monocular video in most cases, \methodname is limited by the ability of the multiview diffusion model, i.e. Era3D~\citep{li2024era3d}, which may have difficulty in handling complex objects and uncommon viewpoints, as shown in Fig.~\ref{fig:limitation}. Improvements in multiview diffusion models could alleviate this problem.

\section{Conclusion}

In this paper, we introduce a new pipeline, called \methodname, to generate 4D videos from just a monocular video. The main challenge in 4D content creation is to simultaneously keep the spatial consistency and the temporal consistency in the generations from diffusion models. Our key idea is to adopt the multiview diffusion models to generate multiview consistent images and then apply the 2D flows to guide the generation of images of different frames to improve temporal consistency. We utilize the information redundancy in a coherent video by adopting the 2D flows to reuse tokens from different frames to generate content for a specific frame. This token re-usage greatly improves the coherence and temporal consistency of the generated images from multiview diffusion models. Experiments demonstrate the effectiveness of our design and show improved quality than baseline methods.

\textbf{Acknowledgment: }This work was supported by the National Natural Science Foundation of China(No.62206174) and MoE Key Laboratory of Intelligent Perception and Human-Machine Collaboration (ShanghaiTech University).

\bibliography{iclr2025_conference}

\begin{thebibliography}{70}
\providecommand{\natexlab}[1]{#1}
\providecommand{\url}[1]{\texttt{#1}}
\expandafter\ifx\csname urlstyle\endcsname\relax
  \providecommand{\doi}[1]{doi: #1}\else
  \providecommand{\doi}{doi: \begingroup \urlstyle{rm}\Url}\fi

\bibitem[An et~al.(2023)An, Zhang, Yang, Gupta, Huang, Luo, and Yin]{an2023latentshiftvd}
Jie An, Songyang Zhang, Harry Yang, Sonal Gupta, Jia-Bin Huang, Jiebo Luo, and Xi~Yin.
\newblock Latent-shift: Latent diffusion with temporal shift for efficient text-to-video generation.
\newblock \emph{arXiv preprint arXiv:2304.08477}, 2023.

\bibitem[Bahmani et~al.(2024)Bahmani, Skorokhodov, Rong, Wetzstein, Guibas, Wonka, Tulyakov, Park, Tagliasacchi, and Lindell]{bahmani20244dfy}
Sherwin Bahmani, Ivan Skorokhodov, Victor Rong, Gordon Wetzstein, Leonidas Guibas, Peter Wonka, Sergey Tulyakov, Jeong~Joon Park, Andrea Tagliasacchi, and David~B Lindell.
\newblock 4d-fy: Text-to-4d generation using hybrid score distillation sampling.
\newblock In \emph{Proceedings of the IEEE/CVF Conference on Computer Vision and Pattern Recognition}, pp.\  7996--8006, 2024.

\bibitem[Blattmann et~al.(2023{\natexlab{a}})Blattmann, Dockhorn, Kulal, Mendelevitch, Kilian, Lorenz, Levi, English, Voleti, Letts, et~al.]{blattmann2023stable}
Andreas Blattmann, Tim Dockhorn, Sumith Kulal, Daniel Mendelevitch, Maciej Kilian, Dominik Lorenz, Yam Levi, Zion English, Vikram Voleti, Adam Letts, et~al.
\newblock Stable video diffusion: Scaling latent video diffusion models to large datasets.
\newblock \emph{arXiv preprint arXiv:2311.15127}, 2023{\natexlab{a}}.

\bibitem[Blattmann et~al.(2023{\natexlab{b}})Blattmann, Rombach, Ling, Dockhorn, Kim, Fidler, and Kreis]{blattmann2023alignlatentvd}
Andreas Blattmann, Robin Rombach, Huan Ling, Tim Dockhorn, Seung~Wook Kim, Sanja Fidler, and Karsten Kreis.
\newblock Align your latents: High-resolution video synthesis with latent diffusion models.
\newblock In \emph{Proceedings of the IEEE/CVF Conference on Computer Vision and Pattern Recognition}, pp.\  22563--22575, 2023{\natexlab{b}}.

\bibitem[Chen et~al.(2023)Chen, Chen, Jiao, and Jia]{chen2023fantasia3d}
Rui Chen, Yongwei Chen, Ningxin Jiao, and Kui Jia.
\newblock Fantasia3d: Disentangling geometry and appearance for high-quality text-to-3d content creation.
\newblock In \emph{Proceedings of the IEEE/CVF international conference on computer vision}, pp.\  22246--22256, 2023.

\bibitem[Dai \& Yang(2024)Dai and Yang]{dai2024curriculum}
Qiyuan Dai and Sibei Yang.
\newblock Curriculum point prompting for weakly-supervised referring image segmentation.
\newblock In \emph{Proceedings of the IEEE/CVF Conference on Computer Vision and Pattern Recognition}, pp.\  13711--13722, 2024.

\bibitem[Deitke et~al.(2023)Deitke, Schwenk, Salvador, Weihs, Michel, VanderBilt, Schmidt, Ehsani, Kembhavi, and Farhadi]{deitke2023objaverse}
Matt Deitke, Dustin Schwenk, Jordi Salvador, Luca Weihs, Oscar Michel, Eli VanderBilt, Ludwig Schmidt, Kiana Ehsani, Aniruddha Kembhavi, and Ali Farhadi.
\newblock Objaverse: A universe of annotated 3d objects.
\newblock In \emph{Proceedings of the IEEE/CVF Conference on Computer Vision and Pattern Recognition}, pp.\  13142--13153, 2023.

\bibitem[Fang et~al.(2022)Fang, Yi, Wang, Xie, Zhang, Liu, Nie{\ss}ner, and Tian]{fang2022fastdnerf}
Jiemin Fang, Taoran Yi, Xinggang Wang, Lingxi Xie, Xiaopeng Zhang, Wenyu Liu, Matthias Nie{\ss}ner, and Qi~Tian.
\newblock Fast dynamic radiance fields with time-aware neural voxels.
\newblock In \emph{SIGGRAPH Asia 2022 Conference Papers}, pp.\  1--9, 2022.

\bibitem[Gao et~al.(2021)Gao, Saraf, Kopf, and Huang]{gao2021dynamicviewsynthesis}
Chen Gao, Ayush Saraf, Johannes Kopf, and Jia-Bin Huang.
\newblock Dynamic view synthesis from dynamic monocular video.
\newblock In \emph{Proceedings of the IEEE/CVF International Conference on Computer Vision}, pp.\  5712--5721, 2021.

\bibitem[Ge et~al.(2023)Ge, Nah, Liu, Poon, Tao, Catanzaro, Jacobs, Huang, Liu, and Balaji]{ge2023preservecovd}
Songwei Ge, Seungjun Nah, Guilin Liu, Tyler Poon, Andrew Tao, Bryan Catanzaro, David Jacobs, Jia-Bin Huang, Ming-Yu Liu, and Yogesh Balaji.
\newblock Preserve your own correlation: A noise prior for video diffusion models.
\newblock In \emph{Proceedings of the IEEE/CVF International Conference on Computer Vision}, pp.\  22930--22941, 2023.

\bibitem[Geyer et~al.(2023)Geyer, Bar-Tal, Bagon, and Dekel]{geyer2023tokenflow}
Michal Geyer, Omer Bar-Tal, Shai Bagon, and Tali Dekel.
\newblock Tokenflow: Consistent diffusion features for consistent video editing.
\newblock \emph{arXiv preprint arXiv:2307.10373}, 2023.

\bibitem[Guo et~al.(2024)Guo, Yang, Rao, Liang, Wang, Qiao, Agrawala, Lin, and Dai]{guo2023animatediffvd}
Yuwei Guo, Ceyuan Yang, Anyi Rao, Zhengyang Liang, Yaohui Wang, Yu~Qiao, Maneesh Agrawala, Dahua Lin, and Bo~Dai.
\newblock Animatediff: Animate your personalized text-to-image diffusion models without specific tuning.
\newblock \emph{International Conference on Learning Representations}, 2024.

\bibitem[Huang et~al.(2024{\natexlab{a}})Huang, Feng, Shi, Xu, Yu, and Yang]{huang2024free}
Hanzhuo Huang, Yufan Feng, Cheng Shi, Lan Xu, Jingyi Yu, and Sibei Yang.
\newblock Free-bloom: Zero-shot text-to-video generator with llm director and ldm animator.
\newblock \emph{Advances in Neural Information Processing Systems}, 36, 2024{\natexlab{a}}.

\bibitem[Huang et~al.(2024{\natexlab{b}})Huang, Sun, Yang, Lyu, Cao, and Qi]{huang2024sc}
Yi-Hua Huang, Yang-Tian Sun, Ziyi Yang, Xiaoyang Lyu, Yan-Pei Cao, and Xiaojuan Qi.
\newblock Sc-gs: Sparse-controlled gaussian splatting for editable dynamic scenes.
\newblock In \emph{Proceedings of the IEEE/CVF Conference on Computer Vision and Pattern Recognition}, pp.\  4220--4230, 2024{\natexlab{b}}.

\bibitem[Jiang et~al.(2023)Jiang, Zhang, Gao, Hu, and Yao]{jiang2023consistent4d}
Yanqin Jiang, Li~Zhang, Jin Gao, Weimin Hu, and Yao Yao.
\newblock Consistent4d: Consistent 360 $\{$$\backslash$deg$\}$ dynamic object generation from monocular video.
\newblock \emph{arXiv preprint arXiv:2311.02848}, 2023.

\bibitem[Jiang et~al.(2024)Jiang, Yu, Cao, Wang, Hu, and Gao]{jiang2024animate3d}
Yanqin Jiang, Chaohui Yu, Chenjie Cao, Fan Wang, Weiming Hu, and Jin Gao.
\newblock Animate3d: Animating any 3d model with multi-view video diffusion.
\newblock \emph{arXiv preprint arXiv:2407.11398}, 2024.

\bibitem[Kerbl et~al.(2023)Kerbl, Kopanas, Leimk{\"u}hler, and Drettakis]{kerbl20233dgs}
Bernhard Kerbl, Georgios Kopanas, Thomas Leimk{\"u}hler, and George Drettakis.
\newblock 3d gaussian splatting for real-time radiance field rendering.
\newblock \emph{ACM Transactions on Graphics}, 42\penalty0 (4), July 2023.
\newblock URL \url{https://repo-sam.inria.fr/fungraph/3d-gaussian-splatting/}.

\bibitem[Li et~al.(2024{\natexlab{a}})Li, Zheng, Zhu, Mai, Zhang, Wonka, and Ghanem]{li2024vividzoo}
Bing Li, Cheng Zheng, Wenxuan Zhu, Jinjie Mai, Biao Zhang, Peter Wonka, and Bernard Ghanem.
\newblock Vivid-zoo: Multi-view video generation with diffusion model.
\newblock \emph{arXiv preprint arXiv:2406.08659}, 2024{\natexlab{a}}.

\bibitem[Li et~al.(2024{\natexlab{b}})Li, Liu, Long, Zhang, Lin, Li, Qi, Zhang, Luo, Tan, et~al.]{li2024era3d}
Peng Li, Yuan Liu, Xiaoxiao Long, Feihu Zhang, Cheng Lin, Mengfei Li, Xingqun Qi, Shanghang Zhang, Wenhan Luo, Ping Tan, et~al.
\newblock Era3d: High-resolution multiview diffusion using efficient row-wise attention.
\newblock \emph{arXiv preprint arXiv:2405.11616}, 2024{\natexlab{b}}.

\bibitem[Li et~al.(2022)Li, Slavcheva, Zollhoefer, Green, Lassner, Kim, Schmidt, Lovegrove, Goesele, Newcombe, et~al.]{li2022neural3dvideo}
Tianye Li, Mira Slavcheva, Michael Zollhoefer, Simon Green, Christoph Lassner, Changil Kim, Tanner Schmidt, Steven Lovegrove, Michael Goesele, Richard Newcombe, et~al.
\newblock Neural 3d video synthesis from multi-view video.
\newblock In \emph{Proceedings of the IEEE/CVF Conference on Computer Vision and Pattern Recognition}, pp.\  5521--5531, 2022.

\bibitem[Li et~al.(2024{\natexlab{c}})Li, Ma, Yang, and Yang]{li2024vidtome}
Xirui Li, Chao Ma, Xiaokang Yang, and Ming-Hsuan Yang.
\newblock Vidtome: Video token merging for zero-shot video editing.
\newblock In \emph{Proceedings of the IEEE/CVF Conference on Computer Vision and Pattern Recognition}, pp.\  7486--7495, 2024{\natexlab{c}}.

\bibitem[Liang et~al.(2024)Liang, Yin, Xu, Liang, Wang, Plataniotis, Zhao, and Wei]{liang2024diffusion4d}
Hanwen Liang, Yuyang Yin, Dejia Xu, Hanxue Liang, Zhangyang Wang, Konstantinos~N Plataniotis, Yao Zhao, and Yunchao Wei.
\newblock Diffusion4d: Fast spatial-temporal consistent 4d generation via video diffusion models.
\newblock \emph{arXiv preprint arXiv:2405.16645}, 2024.

\bibitem[Lin et~al.(2023)Lin, Gao, Tang, Takikawa, Zeng, Huang, Kreis, Fidler, Liu, and Lin]{lin2023magic3d}
Chen-Hsuan Lin, Jun Gao, Luming Tang, Towaki Takikawa, Xiaohui Zeng, Xun Huang, Karsten Kreis, Sanja Fidler, Ming-Yu Liu, and Tsung-Yi Lin.
\newblock Magic3d: High-resolution text-to-3d content creation.
\newblock In \emph{Proceedings of the IEEE/CVF Conference on Computer Vision and Pattern Recognition}, pp.\  300--309, 2023.

\bibitem[Ling et~al.(2024)Ling, Kim, Torralba, Fidler, and Kreis]{ling2024alignyourgauss}
Huan Ling, Seung~Wook Kim, Antonio Torralba, Sanja Fidler, and Karsten Kreis.
\newblock Align your gaussians: Text-to-4d with dynamic 3d gaussians and composed diffusion models.
\newblock In \emph{Proceedings of the IEEE/CVF Conference on Computer Vision and Pattern Recognition}, pp.\  8576--8588, 2024.

\bibitem[Liu et~al.(2023{\natexlab{a}})Liu, Wu, Van~Hoorick, Tokmakov, Zakharov, and Vondrick]{liu2023zero123}
Ruoshi Liu, Rundi Wu, Basile Van~Hoorick, Pavel Tokmakov, Sergey Zakharov, and Carl Vondrick.
\newblock Zero-1-to-3: Zero-shot one image to 3d object.
\newblock In \emph{Proceedings of the IEEE/CVF international conference on computer vision}, pp.\  9298--9309, 2023{\natexlab{a}}.

\bibitem[Liu et~al.(2023{\natexlab{b}})Liu, Lin, Zeng, Long, Liu, Komura, and Wang]{liu2023syncdreamer}
Yuan Liu, Cheng Lin, Zijiao Zeng, Xiaoxiao Long, Lingjie Liu, Taku Komura, and Wenping Wang.
\newblock Syncdreamer: Generating multiview-consistent images from a single-view image.
\newblock \emph{arXiv preprint arXiv:2309.03453}, 2023{\natexlab{b}}.

\bibitem[Liu et~al.(2023{\natexlab{c}})Liu, Wang, Lin, Long, Wang, Liu, Komura, and Wang]{liu2023nero}
Yuan Liu, Peng Wang, Cheng Lin, Xiaoxiao Long, Jiepeng Wang, Lingjie Liu, Taku Komura, and Wenping Wang.
\newblock Nero: Neural geometry and brdf reconstruction of reflective objects from multiview images.
\newblock In \emph{SIGGRAPH}, 2023{\natexlab{c}}.

\bibitem[Long et~al.(2024)Long, Guo, Lin, Liu, Dou, Liu, Ma, Zhang, Habermann, Theobalt, et~al.]{long2024wonder3d}
Xiaoxiao Long, Yuan-Chen Guo, Cheng Lin, Yuan Liu, Zhiyang Dou, Lingjie Liu, Yuexin Ma, Song-Hai Zhang, Marc Habermann, Christian Theobalt, et~al.
\newblock Wonder3d: Single image to 3d using cross-domain diffusion.
\newblock In \emph{Proceedings of the IEEE/CVF Conference on Computer Vision and Pattern Recognition}, pp.\  9970--9980, 2024.

\bibitem[Luiten et~al.(2023)Luiten, Kopanas, Leibe, and Ramanan]{luiten2023dynamic3dgauss}
Jonathon Luiten, Georgios Kopanas, Bastian Leibe, and Deva Ramanan.
\newblock Dynamic 3d gaussians: Tracking by persistent dynamic view synthesis.
\newblock \emph{arXiv preprint arXiv:2308.09713}, 2023.

\bibitem[Mildenhall et~al.(2020)Mildenhall, Srinivasan, Tancik, Barron, Ramamoorthi, and Ng]{mildenhall2020nerf}
Ben Mildenhall, Pratul~P. Srinivasan, Matthew Tancik, Jonathan~T. Barron, Ravi Ramamoorthi, and Ren Ng.
\newblock Nerf: Representing scenes as neural radiance fields for view synthesis.
\newblock In \emph{ECCV}, 2020.

\bibitem[Nichol et~al.(2021)Nichol, Dhariwal, Ramesh, Shyam, Mishkin, McGrew, Sutskever, and Chen]{nichol2021glide}
Alex Nichol, Prafulla Dhariwal, Aditya Ramesh, Pranav Shyam, Pamela Mishkin, Bob McGrew, Ilya Sutskever, and Mark Chen.
\newblock Glide: Towards photorealistic image generation and editing with text-guided diffusion models.
\newblock \emph{arXiv preprint arXiv:2112.10741}, 2021.

\bibitem[Nichol \& Dhariwal(2021)Nichol and Dhariwal]{nichol2021improvedddpm}
Alexander~Quinn Nichol and Prafulla Dhariwal.
\newblock Improved denoising diffusion probabilistic models.
\newblock In \emph{International conference on machine learning}, pp.\  8162--8171. PMLR, 2021.

\bibitem[Pan et~al.(2024)Pan, Yang, Zhu, and Zhang]{pan2024fastdy4d}
Zijie Pan, Zeyu Yang, Xiatian Zhu, and Li~Zhang.
\newblock Fast dynamic 3d object generation from a single-view video.
\newblock \emph{arXiv preprint arXiv:2401.08742}, 2024.

\bibitem[Park et~al.(2021{\natexlab{a}})Park, Sinha, Barron, Bouaziz, Goldman, Seitz, and Martin-Brualla]{park2021nerfies}
Keunhong Park, Utkarsh Sinha, Jonathan~T Barron, Sofien Bouaziz, Dan~B Goldman, Steven~M Seitz, and Ricardo Martin-Brualla.
\newblock Nerfies: Deformable neural radiance fields.
\newblock In \emph{Proceedings of the IEEE/CVF International Conference on Computer Vision}, pp.\  5865--5874, 2021{\natexlab{a}}.

\bibitem[Park et~al.(2021{\natexlab{b}})Park, Sinha, Hedman, Barron, Bouaziz, Goldman, Martin-Brualla, and Seitz]{park2021hypernerf}
Keunhong Park, Utkarsh Sinha, Peter Hedman, Jonathan~T Barron, Sofien Bouaziz, Dan~B Goldman, Ricardo Martin-Brualla, and Steven~M Seitz.
\newblock Hypernerf: A higher-dimensional representation for topologically varying neural radiance fields.
\newblock \emph{arXiv preprint arXiv:2106.13228}, 2021{\natexlab{b}}.

\bibitem[Pons-Moll et~al.(2021)Pons-Moll, Moreno-Noguer, Corona, and Pumarola]{pons2021dnerf}
Gerard Pons-Moll, Francesc Moreno-Noguer, Enric Corona, and Albert Pumarola.
\newblock D-nerf: Neural radiance fields for dynamic scenes.
\newblock In \emph{2021 IEEE/CVF Conference on Computer Vision and Pattern Recognition (CVPR)}. IEEE, 2021.

\bibitem[Poole et~al.(2022)Poole, Jain, Barron, and Mildenhall]{poole2022dreamfusion}
Ben Poole, Ajay Jain, Jonathan~T Barron, and Ben Mildenhall.
\newblock Dreamfusion: Text-to-3d using 2d diffusion.
\newblock \emph{arXiv preprint arXiv:2209.14988}, 2022.

\bibitem[Ramesh et~al.(2022)Ramesh, Dhariwal, Nichol, Chu, and Chen]{ramesh2022hierarchicaldiff}
Aditya Ramesh, Prafulla Dhariwal, Alex Nichol, Casey Chu, and Mark Chen.
\newblock Hierarchical text-conditional image generation with clip latents.
\newblock \emph{arXiv preprint arXiv:2204.06125}, 1\penalty0 (2):\penalty0 3, 2022.

\bibitem[Ren et~al.(2024)Ren, Xie, Mirzaei, Liang, Zeng, Kreis, Liu, Torralba, Fidler, Kim, et~al.]{ren2024l4gm}
Jiawei Ren, Kevin Xie, Ashkan Mirzaei, Hanxue Liang, Xiaohui Zeng, Karsten Kreis, Ziwei Liu, Antonio Torralba, Sanja Fidler, Seung~Wook Kim, et~al.
\newblock L4gm: Large 4d gaussian reconstruction model.
\newblock \emph{arXiv preprint arXiv:2406.10324}, 2024.

\bibitem[Rombach et~al.(2022)Rombach, Blattmann, Lorenz, Esser, and Ommer]{rombach2022latentdiff}
Robin Rombach, Andreas Blattmann, Dominik Lorenz, Patrick Esser, and Bj{\"o}rn Ommer.
\newblock High-resolution image synthesis with latent diffusion models.
\newblock In \emph{Proceedings of the IEEE/CVF conference on computer vision and pattern recognition}, pp.\  10684--10695, 2022.

\bibitem[Schuhmann et~al.(2022)Schuhmann, Beaumont, Vencu, Gordon, Wightman, Cherti, Coombes, Katta, Mullis, Wortsman, et~al.]{schuhmann2022laion}
Christoph Schuhmann, Romain Beaumont, Richard Vencu, Cade Gordon, Ross Wightman, Mehdi Cherti, Theo Coombes, Aarush Katta, Clayton Mullis, Mitchell Wortsman, et~al.
\newblock Laion-5b: An open large-scale dataset for training next generation image-text models.
\newblock \emph{Advances in Neural Information Processing Systems}, 35:\penalty0 25278--25294, 2022.

\bibitem[Shi \& Yang(2022)Shi and Yang]{shi2022spatial}
Cheng Shi and Sibei Yang.
\newblock Spatial and visual perspective-taking via view rotation and relation reasoning for embodied reference understanding.
\newblock In \emph{European Conference on Computer Vision}, pp.\  201--218. Springer, 2022.

\bibitem[Shi \& Yang(2023{\natexlab{a}})Shi and Yang]{shi2023edadet}
Cheng Shi and Sibei Yang.
\newblock Edadet: Open-vocabulary object detection using early dense alignment.
\newblock In \emph{Proceedings of the IEEE/CVF international conference on computer vision}, pp.\  15724--15734, 2023{\natexlab{a}}.

\bibitem[Shi \& Yang(2023{\natexlab{b}})Shi and Yang]{shi2023logoprompt}
Cheng Shi and Sibei Yang.
\newblock Logoprompt: Synthetic text images can be good visual prompts for vision-language models.
\newblock In \emph{Proceedings of the IEEE/CVF International Conference on Computer Vision}, pp.\  2932--2941, 2023{\natexlab{b}}.

\bibitem[Shi \& Yang(2024)Shi and Yang]{shi2024devil}
Cheng Shi and Sibei Yang.
\newblock The devil is in the object boundary: towards annotation-free instance segmentation using foundation models.
\newblock \emph{arXiv preprint arXiv:2404.11957}, 2024.

\bibitem[Shi et~al.(2024{\natexlab{a}})Shi, Zhang, Yang, Tang, Ma, and Yang]{shi2024part2object}
Cheng Shi, Yulin Zhang, Bin Yang, Jiajin Tang, Yuexin Ma, and Sibei Yang.
\newblock Part2object: Hierarchical unsupervised 3d instance segmentation.
\newblock In \emph{European Conference on Computer Vision}, pp.\  1--18. Springer, 2024{\natexlab{a}}.

\bibitem[Shi et~al.(2024{\natexlab{b}})Shi, Zhu, and Yang]{shi2024plain}
Cheng Shi, Yuchen Zhu, and Sibei Yang.
\newblock Plain-det: A plain multi-dataset object detector.
\newblock In \emph{European Conference on Computer Vision}, pp.\  210--226. Springer, 2024{\natexlab{b}}.

\bibitem[Shi et~al.(2023)Shi, Wang, Ye, Long, Li, and Yang]{shi2023mvdream}
Yichun Shi, Peng Wang, Jianglong Ye, Mai Long, Kejie Li, and Xiao Yang.
\newblock Mvdream: Multi-view diffusion for 3d generation.
\newblock \emph{arXiv preprint arXiv:2308.16512}, 2023.

\bibitem[Singer et~al.(2022)Singer, Polyak, Hayes, Yin, An, Zhang, Hu, Yang, Ashual, Gafni, et~al.]{singer2022makeavideovd}
Uriel Singer, Adam Polyak, Thomas Hayes, Xi~Yin, Jie An, Songyang Zhang, Qiyuan Hu, Harry Yang, Oron Ashual, Oran Gafni, et~al.
\newblock Make-a-video: Text-to-video generation without text-video data.
\newblock \emph{arXiv preprint arXiv:2209.14792}, 2022.

\bibitem[Singer et~al.(2023{\natexlab{a}})Singer, Sheynin, Polyak, Ashual, Makarov, Kokkinos, Goyal, Vedaldi, Parikh, Johnson, et~al.]{singer2023mav3d}
Uriel Singer, Shelly Sheynin, Adam Polyak, Oron Ashual, Iurii Makarov, Filippos Kokkinos, Naman Goyal, Andrea Vedaldi, Devi Parikh, Justin Johnson, et~al.
\newblock Text-to-4d dynamic scene generation.
\newblock \emph{arXiv preprint arXiv:2301.11280}, 2023{\natexlab{a}}.

\bibitem[Singer et~al.(2023{\natexlab{b}})Singer, Sheynin, Polyak, Ashual, Makarov, Kokkinos, Goyal, Vedaldi, Parikh, Johnson, et~al.]{singer2023text}
Uriel Singer, Shelly Sheynin, Adam Polyak, Oron Ashual, Iurii Makarov, Filippos Kokkinos, Naman Goyal, Andrea Vedaldi, Devi Parikh, Justin Johnson, et~al.
\newblock Text-to-4d dynamic scene generation.
\newblock \emph{arXiv preprint arXiv:2301.11280}, 2023{\natexlab{b}}.

\bibitem[Tang et~al.(2023{\natexlab{a}})Tang, Zheng, Shi, and Yang]{tang2023contrastive}
Jiajin Tang, Ge~Zheng, Cheng Shi, and Sibei Yang.
\newblock Contrastive grouping with transformer for referring image segmentation.
\newblock In \emph{Proceedings of the IEEE/CVF Conference on Computer Vision and Pattern Recognition}, pp.\  23570--23580, 2023{\natexlab{a}}.

\bibitem[Tang et~al.(2023{\natexlab{b}})Tang, Zheng, and Yang]{tang2023temporal}
Jiajin Tang, Ge~Zheng, and Sibei Yang.
\newblock Temporal collection and distribution for referring video object segmentation.
\newblock In \emph{Proceedings of the IEEE/CVF International Conference on Computer Vision}, pp.\  15466--15476, 2023{\natexlab{b}}.

\bibitem[Tang et~al.(2023{\natexlab{c}})Tang, Zheng, Yu, and Yang]{tang2023cotdet}
Jiajin Tang, Ge~Zheng, Jingyi Yu, and Sibei Yang.
\newblock Cotdet: Affordance knowledge prompting for task driven object detection.
\newblock In \emph{Proceedings of the IEEE/CVF International Conference on Computer Vision}, pp.\  3068--3078, 2023{\natexlab{c}}.

\bibitem[Teed \& Deng(2020)Teed and Deng]{teed2020raft}
Zachary Teed and Jia Deng.
\newblock Raft: Recurrent all-pairs field transforms for optical flow.
\newblock In \emph{Computer Vision--ECCV 2020: 16th European Conference, Glasgow, UK, August 23--28, 2020, Proceedings, Part II 16}, pp.\  402--419. Springer, 2020.

\bibitem[Tretschk et~al.(2021)Tretschk, Tewari, Golyanik, Zollh{\"o}fer, Lassner, and Theobalt]{tretschk2021nonrigidnerf}
Edgar Tretschk, Ayush Tewari, Vladislav Golyanik, Michael Zollh{\"o}fer, Christoph Lassner, and Christian Theobalt.
\newblock Non-rigid neural radiance fields: Reconstruction and novel view synthesis of a dynamic scene from monocular video.
\newblock In \emph{Proceedings of the IEEE/CVF International Conference on Computer Vision}, pp.\  12959--12970, 2021.

\bibitem[Wang et~al.(2021)Wang, Liu, Liu, Theobalt, Komura, and Wang]{wang2021neus}
Peng Wang, Lingjie Liu, Yuan Liu, Christian Theobalt, Taku Komura, and Wenping Wang.
\newblock Neus: Learning neural implicit surfaces by volume rendering for multi-view reconstruction.
\newblock \emph{arXiv preprint arXiv:2106.10689}, 2021.

\bibitem[Wang et~al.(2024)Wang, Lu, Wang, Bao, Li, Su, and Zhu]{wang2024prolificdreamer}
Zhengyi Wang, Cheng Lu, Yikai Wang, Fan Bao, Chongxuan Li, Hang Su, and Jun Zhu.
\newblock Prolificdreamer: High-fidelity and diverse text-to-3d generation with variational score distillation.
\newblock \emph{Advances in Neural Information Processing Systems}, 36, 2024.

\bibitem[Wu et~al.(2024{\natexlab{a}})Wu, Yi, Fang, Xie, Zhang, Wei, Liu, Tian, and Wang]{wu20244dgauss}
Guanjun Wu, Taoran Yi, Jiemin Fang, Lingxi Xie, Xiaopeng Zhang, Wei Wei, Wenyu Liu, Qi~Tian, and Xinggang Wang.
\newblock 4d gaussian splatting for real-time dynamic scene rendering.
\newblock In \emph{Proceedings of the IEEE/CVF Conference on Computer Vision and Pattern Recognition}, pp.\  20310--20320, 2024{\natexlab{a}}.

\bibitem[Wu et~al.(2024{\natexlab{b}})Wu, Yu, Jiang, Cao, Wang, and Bai]{wu2024sc4d}
Zijie Wu, Chaohui Yu, Yanqin Jiang, Chenjie Cao, Fan Wang, and Xiang Bai.
\newblock Sc4d: Sparse-controlled video-to-4d generation and motion transfer.
\newblock \emph{arXiv preprint arXiv:2404.03736}, 2024{\natexlab{b}}.

\bibitem[Xian et~al.(2021)Xian, Huang, Kopf, and Kim]{xian2021spacenerfvideo}
Wenqi Xian, Jia-Bin Huang, Johannes Kopf, and Changil Kim.
\newblock Space-time neural irradiance fields for free-viewpoint video.
\newblock In \emph{Proceedings of the IEEE/CVF conference on computer vision and pattern recognition}, pp.\  9421--9431, 2021.

\bibitem[Yang et~al.(2024{\natexlab{a}})Yang, Pan, Gu, and Zhang]{yang2024diffusion}
Zeyu Yang, Zijie Pan, Chun Gu, and Li~Zhang.
\newblock Diffusion²: Dynamic 3d content generation via score composition of video and multi-view diffusion models.
\newblock \emph{arXiv preprint 2404.02148}, 2024{\natexlab{a}}.

\bibitem[Yang et~al.(2024{\natexlab{b}})Yang, Gao, Zhou, Jiao, Zhang, and Jin]{yang2024deformable4dgauss}
Ziyi Yang, Xinyu Gao, Wen Zhou, Shaohui Jiao, Yuqing Zhang, and Xiaogang Jin.
\newblock Deformable 3d gaussians for high-fidelity monocular dynamic scene reconstruction.
\newblock In \emph{Proceedings of the IEEE/CVF Conference on Computer Vision and Pattern Recognition}, pp.\  20331--20341, 2024{\natexlab{b}}.

\bibitem[Yin et~al.(2023)Yin, Xu, Wang, Zhao, and Wei]{yin20234dgen}
Yuyang Yin, Dejia Xu, Zhangyang Wang, Yao Zhao, and Yunchao Wei.
\newblock 4dgen: Grounded 4d content generation with spatial-temporal consistency.
\newblock \emph{arXiv preprint arXiv:2312.17225}, 2023.

\bibitem[Yuan et~al.(2021)Yuan, Lv, Schmidt, and Lovegrove]{yuan2021stardnerf}
Wentao Yuan, Zhaoyang Lv, Tanner Schmidt, and Steven Lovegrove.
\newblock Star: Self-supervised tracking and reconstruction of rigid objects in motion with neural rendering.
\newblock In \emph{Proceedings of the IEEE/CVF Conference on Computer Vision and Pattern Recognition}, pp.\  13144--13152, 2021.

\bibitem[Zeng et~al.(2024)Zeng, Jiang, Zhu, Lu, Lin, Zhu, Hu, Cao, and Yao]{zeng2024stag4d}
Yifei Zeng, Yanqin Jiang, Siyu Zhu, Yuanxun Lu, Youtian Lin, Hao Zhu, Weiming Hu, Xun Cao, and Yao Yao.
\newblock Stag4d: Spatial-temporal anchored generative 4d gaussians.
\newblock \emph{arXiv preprint arXiv:2403.14939}, 2024.

\bibitem[Zhang et~al.(2024)Zhang, Chen, Wang, Liu, Wang, and Qiao]{zhang20244diffusion}
Haiyu Zhang, Xinyuan Chen, Yaohui Wang, Xihui Liu, Yunhong Wang, and Yu~Qiao.
\newblock 4diffusion: Multi-view video diffusion model for 4d generation.
\newblock \emph{arXiv preprint arXiv:2405.20674}, 2024.

\bibitem[Zhao et~al.(2023)Zhao, Yan, Xie, Hong, Li, and Lee]{zhao2023animate124}
Yuyang Zhao, Zhiwen Yan, Enze Xie, Lanqing Hong, Zhenguo Li, and Gim~Hee Lee.
\newblock Animate124: Animating one image to 4d dynamic scene.
\newblock \emph{arXiv preprint arXiv:2311.14603}, 2023.

\bibitem[Zheng et~al.(2023)Zheng, Yang, Tang, Zhou, and Yang]{zheng2023ddcot}
Ge~Zheng, Bin Yang, Jiajin Tang, Hong-Yu Zhou, and Sibei Yang.
\newblock Ddcot: Duty-distinct chain-of-thought prompting for multimodal reasoning in language models.
\newblock \emph{Advances in Neural Information Processing Systems}, 36:\penalty0 5168--5191, 2023.

\bibitem[Zheng et~al.(2024)Zheng, Li, Nagano, Liu, Hilliges, and De~Mello]{zheng2024dreamin4d}
Yufeng Zheng, Xueting Li, Koki Nagano, Sifei Liu, Otmar Hilliges, and Shalini De~Mello.
\newblock A unified approach for text-and image-guided 4d scene generation.
\newblock In \emph{Proceedings of the IEEE/CVF Conference on Computer Vision and Pattern Recognition}, pp.\  7300--7309, 2024.

\end{thebibliography}
\bibliographystyle{iclr2025_conference}

\appendix

\newpage
\section{Appendix}

\subsection{Additional Implementation Details}

\textbf{Loss functions.} As introduced in Sec.~\ref{sec:recon}, we employed several losses during the training of the dynamic Gaussian field, including rendering loss, mask loss, DSSIM loss, ARAP loss, normal map loss, and 2D flow loss. We denote these loss as $L_r$, $L_m$, $L_{DSSIM}$, $L_{arap}$, $L_n$, and $L_f$, respectively. Thus, our loss function can be expressed as $L = \lambda_r L_r +\lambda_m L_m + \lambda_{DSSIM} L_{DSSIM} +\lambda_{arap} L_{arap} + \lambda_n L_n + \lambda_f L_f$, where $\lambda$ represent hyperparameters. For general cases, we set $\lambda_r$ to 0.8, $\lambda_{DSSIM}$ to 0.2, $\lambda_m$ to 2, and the remaining hyperparameters to 1.

\textbf{Training iterations.} Following~\citep{kerbl20233dgs}, our training consists of a total of 30K iterations. We first use 5K iterations to learn a static 3D Gaussian from multiview images of a keyframe, which serves as the initialization for the dynamic 3D Gaussian representation. Next, we utilize 10K iterations to learn a coarse dynamic 3D Gaussian field from multiview videos. After regenerating the multiview videos with improved quality, we perform 15K iterations to refine and obtain the final dynamic 3D Gaussian field.
The diffusion process requires approximately 30 GB of GPU memory, while the Gaussian field reconstruction utilizes around 10 GB. The comparison of training times with other methods is shown in Table~\ref{tab:time}.
\begin{table*}[h]
\caption{Training time Comparison with other methods. The number of other methods presented in the table is sourced from~\cite{yang2024diffusion}.}
\vspace{1pt}
\setlength{\tabcolsep}{0.25em} %
\centering
\begin{tabular}{lcccccc}
\toprule
 & Consistent4D & SC4D & Stag4D & DG4D & $\text{Diffusion}^2$ & Ours    \\
\midrule
Time & 120 mins & 30 mins & 90 mins  & 11 mins & 12 mins & 120 mins \\
\bottomrule
\end{tabular}%
\label{tab:time}
\end{table*}

\textbf{Sampling strategy for training the dynamic 3D Gaussian field.} During each training step of the dynamic 3D Gaussian field, we sample various timesteps and utilize all corresponding multiview images for training. Specifically, we employ paired flow inputs to compute the flow loss. However, when the intervals between sampled timesteps are large, rapid movements or changes increase the risk of unreliable flow, which is detrimental to training the dynamic 3D Gaussian field. To mitigate this issue, we implemented an imbalanced sampling strategy that increases the probability of sampling adjacent frames simultaneously.

\begin{figure}[b]
    \centering
    \includegraphics[width=0.98\linewidth]{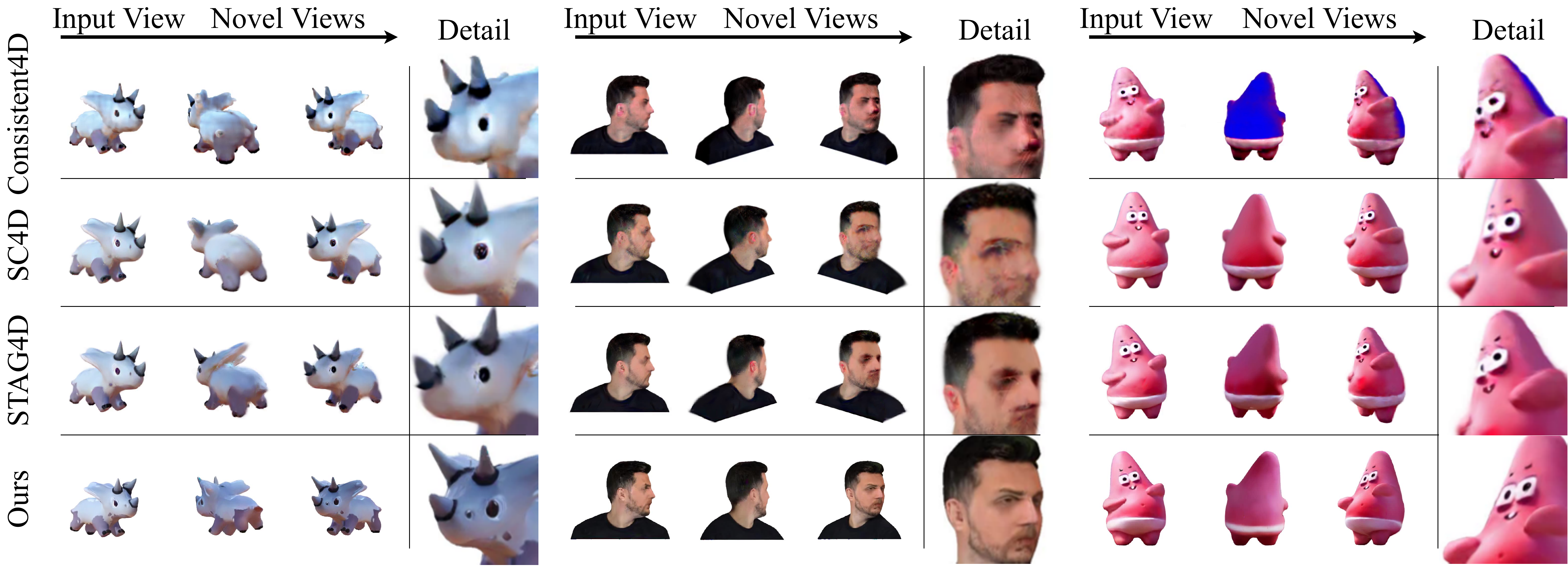}
    \caption{Additional results on spatial consistency with baseline methods, Consistent4D~\citep{jiang2023consistent4d}, SC4D~\citep{wu2024sc4d}, and STAG4D~\citep{zeng2024stag4d}.}
    \label{fig:compare_view2}
\end{figure}

\subsection{Custom Data Preparation}

We manually select high-quality clips from web-collected video data, ensuring each clip contains fewer than 64 frames. Dynamic objects of interest are segmented using SAM2, after which we apply padding to the segmented objects before inputting them into MVD. The resulting videos typically span approximately 4 to 5 seconds.

\subsection{Additional Results}
\label{sec:addtional-results}
\begin{figure}[t]
    \centering
    \includegraphics[width=0.98\linewidth]{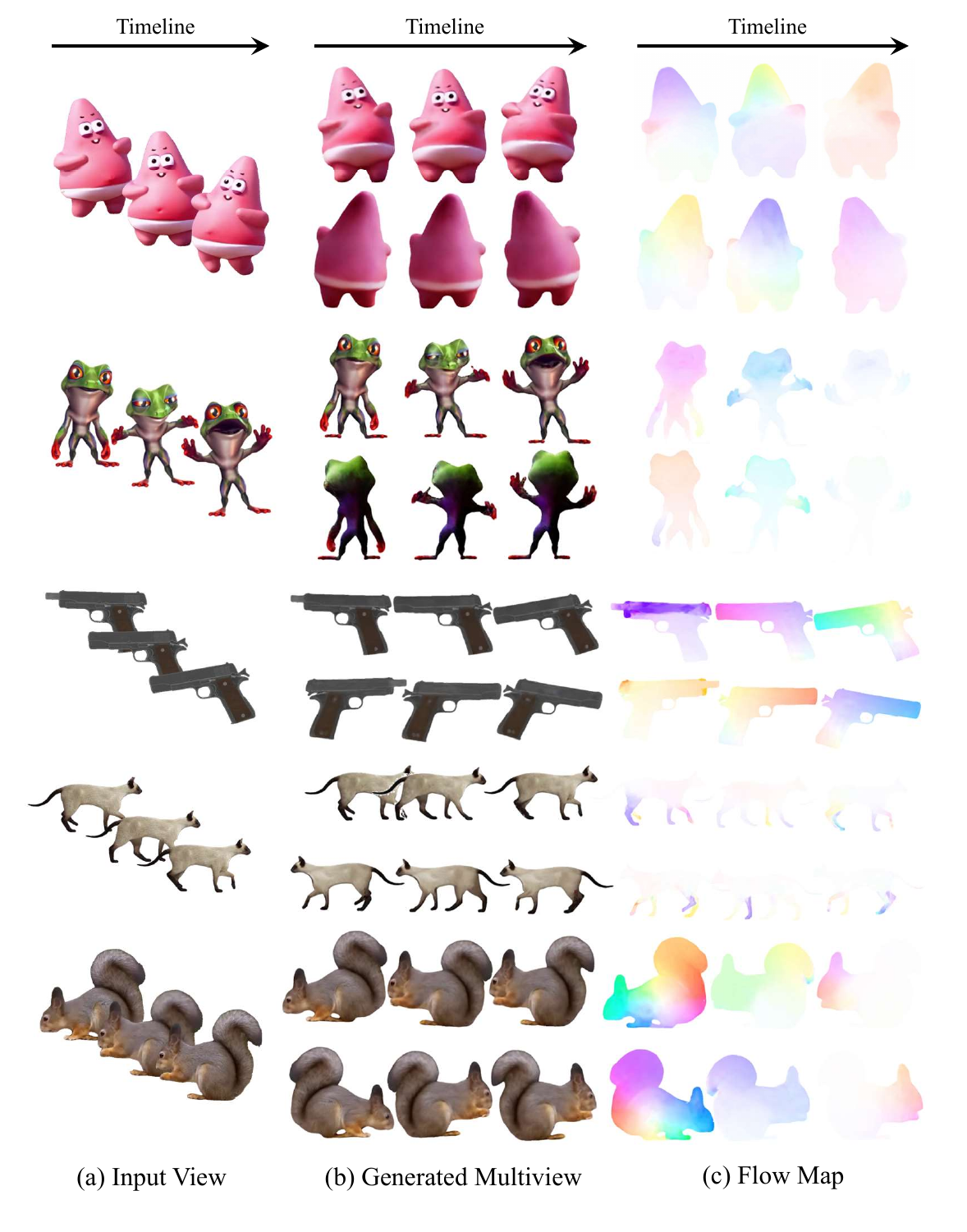}
    \caption{Additional results with flow map of samples in Consistent4D.}
    \label{fig:add_in}
\end{figure}

\begin{figure}[t]
    \centering
    \includegraphics[width=0.98\linewidth]{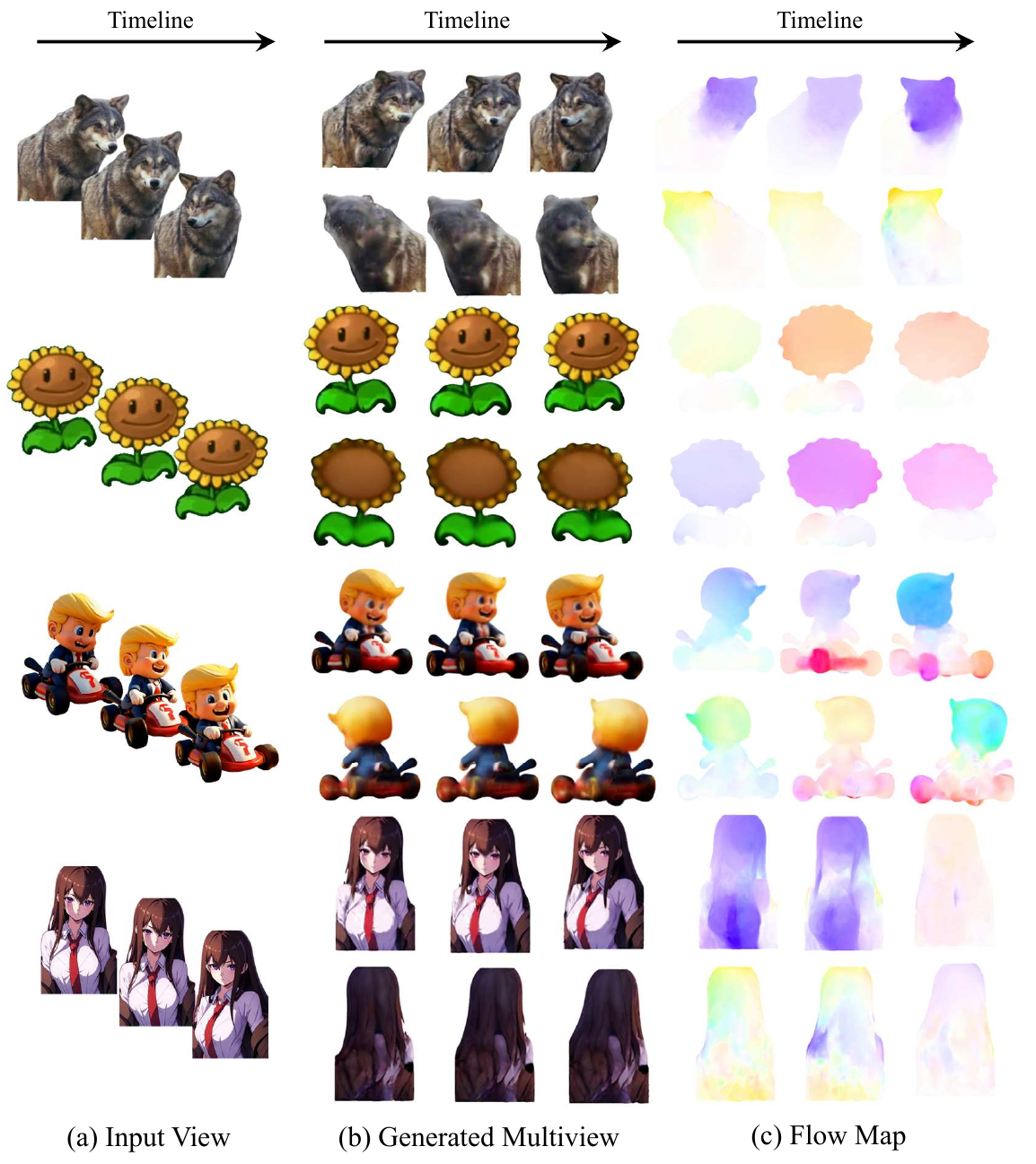}
    \caption{Additional results with flow maps of samples collected on the Internet.}
    \label{fig:add_out_1}
\end{figure}

\begin{figure}[htbp!]
    \centering
    \includegraphics[width=0.85\linewidth]{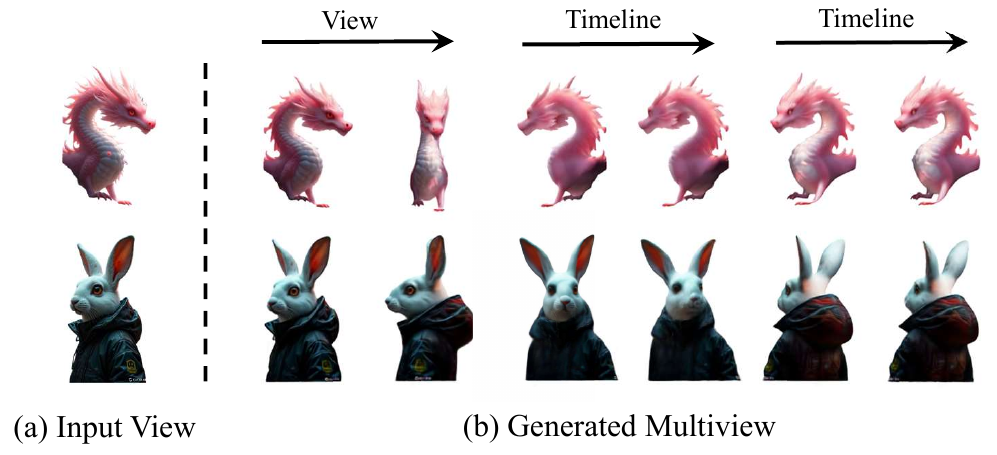}
    \caption{Additional results with more viewpoints of samples collected on the Internet.}
    \label{fig:add_out_2}
\end{figure}

We show additional results on the Consistent4D dataset and a self-collected dataset including the novel-view images and the rendered 2D flow maps in Fig.~\ref{fig:add_in} and Fig.~\ref{fig:add_out_1}. In addition, results with more viewpoints are presented in Fig.~\ref{fig:compare_view2} and Fig.~\ref{fig:add_out_2}. For the self-collected dataset, we use videos collected from the Internet.

\end{document}